%% file: main.tex
\documentclass[twoside]{article}

\usepackage[accepted]{aistats2020}
\input{math_commands.tex}

\usepackage{todonotes}
\usepackage{microtype}
\usepackage{graphicx}
\usepackage{multirow}
\usepackage{subcaption}
\usepackage{booktabs} %

\usepackage{enumitem}
\setlist{nolistsep}
\usepackage{hyperref}
\usepackage{amssymb}

\usepackage{float}
\usepackage{url}
\usepackage{color}

\definecolor{airforceblue}{rgb}{0.36, 0.54, 0.66}
\definecolor{amaranth}{rgb}{0.9, 0.17, 0.31}
\definecolor{cerulean}{rgb}{0.0, 0.48, 0.65}
\definecolor{burntorange}{rgb}{0.8, 0.33, 0.0}
\definecolor{darkorchid}{rgb}{0.6, 0.2, 0.8}
\definecolor{darkterracotta}{rgb}{0.8, 0.31, 0.36}
\definecolor{dartmouthgreen}{rgb}{0.05, 0.5, 0.06}
\definecolor{darkpastelred}{rgb}{0.76, 0.23, 0.13}

\newtheorem{proposition}{Proposition}
\newtheorem{example}{Example}

\definecolor{dartmouthgreen}{rgb}{0.05, 0.5, 0.06}

\newcommand{\q}{q_\theta}

\newcommand{\thetamle}{\theta^{\text{MLE}}}
\usepackage{enumitem}

\definecolor{mygreen}{rgb}{0.05, 0.5, 0.06}

\usepackage[round]{natbib}

\newcommand{\appropto}{\mathrel{\vcenter{
  \offinterlineskip\halign{\hfil$##$\cr
    \propto\cr\noalign{\kern2pt}\sim\cr\noalign{\kern-2pt}}}}}

\begin{document}

\runningtitle{On the interplay between noise and curvature}

\runningauthor{Thomas, Pedregosa, van Merri\"{e}nboer, Manzagol, Bengio, Le Roux}

\twocolumn[

\aistatstitle{On the interplay between noise and curvature\\ and its effect on optimization and generalization}

\aistatsauthor{ Valentin Thomas \And Fabian Pedregosa \And Bart van Merri\"{e}nboer} 

\aistatsaddress{Mila, Universit\'{e} de Montr\'{e}al \And Google Research, Brain Team \And Google Research, Brain Team} 

\aistatsauthor{ Pierre-Antoine Manzagol \And Yoshua Bengio \And Nicolas Le Roux }

\aistatsaddress{ Google Research, Brain Team \And Mila, Universit\'{e} de Montr\'{e}al\\ CIFAR Senior Fellow \And Google Research, Brain Team\\ Mila, McGill University} 
]

\begin{abstract}
    The speed at which one can minimize an expected loss using stochastic methods depends on two properties: the curvature of the loss and the variance of the gradients. While most previous works focus on one or the other of these properties, we explore how their interaction affects optimization speed. Further, as the ultimate goal is good generalization performance, we clarify how both curvature and noise are relevant to properly estimate the generalization gap. Realizing that the limitations of some existing works stems from a confusion between these matrices, we also clarify the distinction between the Fisher matrix, the Hessian, and the covariance matrix of the gradients.
\end{abstract}

\setcounter{footnote}{0}

\section{Introduction}

Training a machine learning model is often cast as the minimization of a smooth function $f$ over parameters $\theta$ in $\mathbb{R}^d$. More precisely, we aim at finding a minimum of an expected loss, i.e.
\begin{align}
    \theta^\ast &\in \arg\min_\theta \E_{p} [\ell(\theta, x)] \; ,
\end{align}
where the expectation is under the data distribution $x \sim p$. In practice, we only have access to an empirical distribution $\hat{p}$ over $x$ and minimize the training loss
\begin{align}
    \hat{\theta}^\ast   &\in \arg\min_\theta \E_{\hat{p}}[\ell(\theta, x)]\\
                        &= \arg\min_\theta f(\theta) \; .
\end{align}
To minimize this function, we assume access to an oracle which, for every value of $\theta$ and $x$, returns both $\ell(\theta, x)$ and its derivative with respect to $\theta$, i.e., $\nabla_\theta \ell(\theta, x)$. Given this oracle, stochastic gradient iteratively performs the following update: $\theta_{t+1} = \theta_t - \alpha_t \nabla \ell(\theta_t, x)$~\footnote{We omit the subscripts when clear from context.} where $\{\alpha_t\}_{t \geq 0}$ is a sequence of stepsizes.

Two questions arise: First, how quickly do we converge to $\hat{\theta}^\ast$ and how is this speed affected by properties of $\ell$ and $\hat{p}$? Second, what is $\E_{p}[\ell(\hat{\theta}^\ast, x)]$?

It is known that the former is influenced by two quantities: the curvature of the function, either measured through its smoothness constant or its condition number, and the noise on the gradients, usually measured through a bound on $\E_{\hat{p}}[\|\nabla \ell(\theta, x)\|^2]$. For instance,  when $f$ is $\mu$-strongly convex, $L$-smooth, and the noise is bounded, i.e. $\E_{\hat{p}}[\|\nabla\ell(\theta, x)\|^2] \leq c$, then stochastic gradient with a constant stepsize $\alpha$ will converge linearly to a ball~\citep{schmidt2014convergence}. Calling $\Delta$ the suboptimality, i.e. $\Delta_k = f(\theta_k) - f(\hat{\theta}^\ast)$, we have
\begin{align}
    \label{eq:sg_cvg_rate}
    \E [\Delta_k]   &\leq (1 - 2\alpha\mu)^k \Delta_0 + \frac{L\alpha c}{4\mu} \; .
\end{align}

This implies that as $k\to \infty$, the expected suboptimality depends on both the curvature through $\mu$ and $L$, and on the noise through $c$. However, bounding curvature and noise using constants rather than full matrices hides the dependencies between these two quantities. We also observed that, because existing works replace the full noise matrix with a constant when deriving convergence rates, that matrix is poorly understood and is often confused with a curvature matrix.
This confusion remains when discussing the generalization properties of a model. Indeed, the generalization gap stems from a discrepancy between the empirical and the true data distribution. An estimator of this gap must thus include an estimate of that discrepancy in addition to an estimate of the impact of an infinitesimal discrepancy on the loss. The former can be characterized as noise and the latter as curvature. Hence, attempts at estimating the generalization gap using only the curvature~\citep{keskar2016large,novak2018sensitivity} are bound to fail as do not characterize the size or geometry of the discrepancy.

In this work, we make the following contributions:
\begin{itemize}
    \item We provide theoretical and empirical evidence of the similarities and differences surrounding the curvatures matrices; the Fisher $\rmF$ and the Hessian $\rmH$, and the noise matrix, $\rmC$;
    \item We briefly expand the convergence results of~\citet{schmidt2014convergence}, theoretically and empirically highlighting the importance of the relationship between noise and curvature for strongly convex functions and quadratics;
    \item We make the connection with an old estimator of the generalization gap, the Takeuchi Information Criterion, and show how its use of both curvature and noise yields a superior estimator to other commonly used ones, such as flatness or sensitivity for neural networks.
\end{itemize}

\section{Information matrices: definitions, similarities, and differences}
\label{sec:hfc}
Before delving into the impact of the information matrices for optimization and generalization, we start by recalling their definitions. We shall see that, despite having similar formulations, they encode different information. We then provide insights on their similarities and differences.

We discuss here two information matrices associated with curvature, the Fisher matrix $\rmF$ and the Hessian $\rmH$, and one associated with noise, the gradients' uncentered covariance $\rmC$. In particular, while $\rmF$ and $\rmH$ are well understood, $\rmC$ is often misinterpreted. For instance, it is often called ``empirical Fisher''~\citep{martens2014new} despite bearing no relationship to $\rmF$, the true Fisher. This confusion can have dire consequences and optimizers using $\rmC$ as approximation to $\rmF$ can have arbitrarily poor performance~\citep{kunstner2019limitations}.

To present these matrices, we consider the case of maximum likelihood estimation (MLE). We have access to a set of samples $(x,y) \in \gX \times \gY$ where $x$ is the input and $y$ the target. We define $p: \gX \times \gY \mapsto \sR$ as the \textbf{data distribution} and $\q: \gX \times \gY \mapsto \sR$ such that $\q(x,y) = p(x) \q(y|x)$ as the \textbf{model distribution}\footnote{$\q(y|x)$ are the softmax activations of a neural network in the classification setting.}.
For each sample $(x, y) \sim p$, our loss is the negative log-likelihood $\ell(\theta, y,x) = - \log \q(y|x)$.
Note that all the definitions and results in this section are valid whether we use the true data distribution $p$ or the empirical $\hat{p}$.

Matrices $\rmH$, $\rmF$ and $\rmC$ are then defined as:

\begin{align}
\rmH(\theta) &= \E_{\color{darkorchid}{p}}\color{cerulean}{\left[\frac{\partial^2 }{\partial \theta \partial \theta^\top} \ell(\theta, y,x)\right]} \label{eq:def_h}\\
\rmC(\theta) &= \E_{\color{darkorchid}{p}}\color{burntorange}{\left[\frac{\partial}{\partial \theta }\ell(\theta, y,x) \frac{\partial}{\partial \theta }\ell(\theta, y,x)^\top\right]} \label{eq:def_c}\\
\rmF(\theta) &= \E_{\color{dartmouthgreen}{q_\theta}}\color{burntorange}{\left[\frac{\partial}{\partial \theta }\ell(\theta, y,x) \frac{\partial}{\partial \theta }\ell(\theta, y,x)^\top\right]} \label{eq:def_f1}\\
&= \E_{\color{dartmouthgreen}{q_\theta}}\color{cerulean}{\left[\frac{\partial^2 }{\partial \theta \partial \theta^\top} \ell(\theta, y,x)\right]}  \; .\label{eq:def_f2}
\end{align}

We observe the following: a) The definition of $\rmH$ and $\rmC$ involves the data distribution, in contrast with the definition of $\rmF$, which involves the model distribution; b) If $\q = p$, all matrices are equal. Furthermore, as noted by~\citet{martens2014new}, $\rmH = \rmF$ whenever the matrix of second derivatives does not depend on $y$, a property shared in particular by all generalized linear models.

As said above, $\rmH$, $\rmF$, and $\rmC$ characterize different properties of the optimization problem. $\rmH$ and $\rmF$ are curvature matrices and describe the geometry of the space around the current point. $\rmC$, on the other hand, is a ``noise matrix'' and represents the sensitivity of the gradient to the particular sample.\footnote{Technically, it is $\rmS$, the centered covariance matrix, rather than $\rmC$ which plays that role but the two are similar close to a stationary point.}

We now explore in more details their similarities and differences.

\subsection{Bounds between \texorpdfstring{$\rmH$}{H}, \texorpdfstring{$\rmF$}{F} and \texorpdfstring{$\rmC$}{C}}
The following proposition bounds the distance between the information matrices:

\begin{proposition}[Distance between $\rmH, \rmF$ and $\rmC$]
Assuming the second moments of the Fisher are bounded above, i.e. $ \E_{\q} [ ||\nabla_\theta^2 \ell(\theta, x, y)||^2] \leq \beta_1$ and $\E_{\q} [ ||\nabla_\theta \ell(\theta, x, y) \nabla_\theta \ell(\theta, x, y)^\top||^2] \leq \beta_2$, we have
\begin{eqnarray*}
||\rmF - \rmH||^2 &\le& \beta_1 \ \gD_{\chi^2}(p||\q) \; ,\\
||\rmF - \rmC||^2 &\le&  \beta_2 \ \gD_{\chi^2}(p||\q) \; ,\\
||\rmC - \rmH||^2 &\le&  \big(\beta_1 + \beta_2\big) \ \gD_{\chi^2}(p||\q) \; .
\end{eqnarray*}
where $\gD_{\chi^2}(p||\q) = \iint \tfrac{(p(x, y)-\q(x, y))^2}{\q(x, y)} dy dx $ is the $\chi^2$ divergence and $|| \cdot ||$ is the Frobenius norm.
\end{proposition}
All the proofs are in the appendix.

In particular, when $p = \q$ we recover that $\rmF = \rmH = \rmC$. At first glance, one could assume that, as the model is trained and $q_\theta$ approaches $p$, these matrices become more similar. The $\chi^2$ divergence is however poorly correlated with the loss of the model. One sample where the prediction of the model is much smaller than the true distribution can dramatically impact the $\chi^2$ divergence and thus the distance between the information matrices. We show in Section~\ref{sec:HFC_diff} how these distances evolve when training a deep network.

\subsection{\texorpdfstring{$\rmC$}{C} does not approximate \texorpdfstring{$\rmF$}{F}}
\label{sec:C_is_not_F}
$\rmC$ is often referred to as the ``empirical Fisher'' matrix, implying that it is an approximation to the true Fisher matrix $\rmF$~\citep{martens2014new}. In some recent works~\citep{liang2017fisher, george2018fast} the empirical Fisher matrix is used instead of the Fisher matrix.
However, in the general case, there is no guarantee that $\rmC$ will approximate $\rmF$, even in the limit of infinite samples. We now give a simple example highlighting their different roles:
\begin{example}[Mean regression]
Let $X = (x_i)_{i=1, \dots, N}$ be an i.i.d sequence of random variables. The task is to estimate $\mu = \E [x]$ by minimizing the loss $ \ell(\theta) = \frac{1}{2 N} \sum_{n=1}^N||x_n - \theta||^2 $. The minimum is attained at $\thetamle= \frac{1}{N} \sum_{n=1}^N x_n$. This estimator is consistent and converges to $\mu$ at rate $\mathcal{O}(\tfrac{1}{\sqrt{N}})$.

This problem is an MLE problem if we define $\q(x) = \mathcal{N}(x; \theta, \rmI_d)$. In this case, we have
\begin{align}
\label{eq:HFC}
\rmH(\thetamle) = \rmF(\thetamle) = \rmI_d\quad &, \quad \rmC(\thetamle) = \widehat{\rmSS}_x \; ,
\end{align}
where $\widehat{\rmSS}_x$ is the empirical covariance of the $x_i$'s. We see that, even in the limit of infinite data, the covariance $\rmC$ does not converge to the actual Fisher matrix nor the Hessian. Hence we shall and will not refer to $\rmC$ as the ``empirical Fisher'' matrix.
\end{example}

In some other settings, however, one can expect a stronger correlation between $\rmC$ and $\rmH$:
\begin{example}(Ordinary Least Squares)
Let us assume we have a data distribution $(x_n, y_n)$ so that
\begin{equation}\label{eq:linear_model}
    y_n = x_n^\top \theta^\ast  + \epsilon_n, \quad \epsilon_n \sim \mathcal{N}(0, \rmSS)
\end{equation}
with $y_n \in \sR^p$ and $x_n, \epsilon_n \in \sR^d$, $\theta \in \sR^{p \times d}$.
We train a model to minimize the sum of squares residual
\begin{equation}
\min_\theta\frac{1}{2 N} \sum_{n=1}^N || y_n - x_n^\top\theta ||^2 \; .
\end{equation}
In that case, for $\theta = \theta^\ast$
\begin{align}
\label{eq:HFC_OLS}
\rmH = \rmF = \E_p [ x x^\top ] \quad &, \quad \rmC =\E_p [ x \rmSS x^\top ]  \; 
\end{align}
First, we observe that the Hessian and the Fisher are equal, for all $\theta$. Second, when the input/output covariance matrix is isotropic $\rmSS=\sigma^2 \rmI$, then we have $\rmC \propto \rmH = \rmF$. 
\end{example}

\section{Information matrices in optimization}
\label{sec:optimization}
Now that the distinction between these matrices has been clarified, we can explain how their interplay affects optimization. In this section, we offer theoretical results, as well as a small empirical study, on the impact of the noise geometry on convergence rates.

\subsection{Convergence rates}
We start here by expanding the the result of~\citet{schmidt2014convergence} to full matrices, expliciting how the interplay of noise and curvature affects optimization. We then build a toy experiment to validate the results.

\subsubsection{General setting}

\begin{proposition}[Function value]
\label{prop:function_value}
Let $f$ be a twice-differentiable function. Assume $f$ is $\mu$-strongly convex and there are two matrices $\rmH$ and $\rmS$ such that for all $\theta$, $\theta'$:
\begin{align*}
    f(\theta') &\leq f(\theta) + \nabla f(\theta)^\top(\theta' - \theta) + \frac{1}{2} (\theta' - \theta)^\top \rmH (\theta' - \theta)\\
    &\E_p[\nabla \ell (\theta, x) \nabla \ell (\theta, x)^\top] \preccurlyeq \rmS + \nabla f(\theta) \nabla f(\theta)^\top\; .
\end{align*}
Then stochastic gradient with stepsize $\alpha$ and positive definite preconditioning matrix $\rmM$ satisfies
\begin{align*}
    \E [\Delta_k]   &\leq (1 - 2\alpha\mu_M\mu)^k \Delta_0 + \frac{\alpha}{4\mu_{M} \mu }\Tr(\rmH \rmM \rmS \rmM) \; ,
\end{align*}
where $\mu_M$ is the smallest eigenvalue of\, $\rmM - \frac{\alpha}{2} \rmM^\top \rmH \rmM$.
\end{proposition}

$\rmS$ is the centered covariance matrix of the stochastic gradients and, if $f$ is quadratic, then $\rmH$ is the Hessian.

\subsubsection{Centered and uncentered covariance}
Proposition~\ref{prop:function_value}, as well as most results on optimization, uses a bound on the uncentered covariance of the gradients. The result is that the noise must be lower far away from the optimum, where the gradients are high. Thus, it seems more natural to define convergence rates as a function of the \emph{centered} gradients' covariance $\rmS$, although these results are usually weaker as a consequence of the relaxed assumption. For the remainder of this section, focused on the quadratic case, we will use $\rmS$. Note that the two matrices are equal at any first-order stationary point.

Centered covariance matrices have been used in the past to derive convergence rates, for instance by~\citet{bach2013non,flammarion2015averaging,dieuleveut2016nonparametric}. These works also include a dependence on the geometry of the noise since their constraint is of the form $\rmS \preccurlyeq \sigma^2 \rmH$. In particular, if $\rmS$ and $\rmH$ are not aligned, $\sigma^2$ must be larger for the inequality to hold.

\subsubsection{Quadratic functions}
\begin{proposition}[Quadratic case]
\label{prop:quadratic}
Assuming we minimize a quadratic function $$f(\theta) = \tfrac{1}{2} (\theta - \hat\theta^\ast)^\top \rmH (\theta-\hat\theta^\ast)$$ only having access to a noisy estimate of the gradient $g(\theta) \sim \nabla f(\theta) + \epsilon$ with $\epsilon$ a zero-mean random variable with covariance $\E [\epsilon \epsilon^\top ] = \rmS$, then the iterates obtained using stochastic gradient with stepsize $\alpha$ and preconditioning psd matrix $\rmM$ satisfy
\begin{align*}
    \E [\theta^k - \hat\theta^\ast] &= (1 - \alpha \rmM \rmH)^k (\theta^0 - \hat\theta^\ast) \; .
\end{align*}
Further, the covariance of the iterates $\rmSS_k = {\E [ (\theta^k-\hat\theta^\ast) (\theta^k-\hat\theta^\ast)^\top]} $ satisfies
\begin{align*}
    \rmSS_{k+1} &=  (\rmI  -\alpha \rmM \rmH) \rmSS_{k}  (\rmI  -\alpha \rmM \rmH)^\top + \alpha^2 \rmM \rmS \rmM^\top\; .
\end{align*}
In particular, the stationary distribution of the iterates has a covariance $\rmSS_\infty$ which verifies the equation
\begin{align*}
    \rmSS_\infty \rmH \rmM + \rmM \rmH \rmSS_\infty &=  \alpha \rmM \big( \rmS + \rmH \rmSS_\infty \rmH\big) \rmM\; .
\end{align*}
\end{proposition}
Recently, analyzing the stationary distribution of SGD by modelling its dynamics as a stochastic differential equation (SDE) has gained traction in the machine learning community~\citep{chaudhari2017stochastic, jastrzkebski2017three}. Worthy of note, \citet{mandt2017stochastic, zhu2018anisotropic} do not make assumptions about the structure of the noise matrix $\rmS$. Our proposition above extends and corrects some of their results as it does not rely on the continuous-time approximation of the dynamics of SGD. Indeed, as pointed out in~\citet{yaida2018fluctuation}, most of the works using the continuous-time approximation implicitly make the confusion between centered $\rmS$ and uncentered $\rmC$ covariance matrices of the gradients.

\begin{proposition}[Limit cycle of SG]
\label{prop:sg_limit} 
If $f$ is a quadratic function and $\rmH$, $\rmC$ and $\rmM$ are simultaneously diagonalizable, then stochastic gradient with symmetric positive definite preconditioner $\rmM$ and stepsize $\alpha$ yields
\begin{align}
\E[\Delta_t]  &= \tfrac{\alpha}{2} \Tr ((2 \rmI - \alpha \rmM \rmH)^{-1} \rmM \rmS) + \gO(e^{-t}) \; .
\end{align}
\end{proposition}

Rather than using a preconditioner, another popular method to reduce the impact of the curvature is Polyak momentum, defined as
\begin{align*}
    v_0 &=0 \quad , \quad v_t = \gamma v_{t-1} + \nabla_\theta \ell(\theta_t, x_t)\\
    \theta_{t+1}    &= \theta_t - \alpha v_t \; .
\end{align*}

\begin{proposition}[Limit cycle of momentum]
\label{prop:polyak_limit}
If $f$ is a quadratic function and $\rmH$ and $\rmC$ are simultaneously diagonalizable, then Polyak momentum with parameter $\gamma$ and stepsize $\alpha$ yields
\begin{align}
    \E[\Delta_t]    &= \tfrac{\alpha}{2} \tfrac{(1 + \gamma)}{(1 - \gamma)} \Tr \left((2 (1 + \gamma) \rmI - \alpha \rmH)^{-1} \rmS \right) + \gO(e^{-t}) \; .
\end{align}
\end{proposition}

\section{Generalization}
So far, we focused on the impact of the interplay between curvature and noise in the optimization setting. However, optimization, i.e. reaching low loss on the training set, is generally not the ultimate goal as one would rather reach a low test loss. The difference between training and test loss is called the generalization gap and estimating it has been the focus of many authors~\citep{keskar2016large, neyshabur2017exploring, liang2017fisher, novak2018sensitivity, rangamani2019scale}.

We believe there is a fundamental misunderstanding in several of these works, stemming from the confusion between curvature and noise. Rather than proposing a new metric, we empirically show how the Takeuchi information criterion~\citep[TIC:][]{takeuchi1976distribution} addresses these misunderstandings. It makes use of both the Hessian of the loss with respect to the parameters, $\rmH$, and the uncentered covariance of the gradients, $\rmC$. While the former represents the curvature of the loss, i.e., the sensitivity of the gradient to a change in parameter space, the latter represents the sensitivity of the gradient to a change in inputs. As the generalization gap is a direct consequence of the discrepancy between training and test sets, the influence of $\rmC$ is natural. Thus, our result further reinforces the idea that the Hessian cannot by itself be used to estimate the generalization gap, an observation already made by~\citet{dinh2017sharp}, among others.

\subsection{Takeuchi information criterion}
In the simplest case of a well specified least squares regression problem, an unbiased estimator of the generalization gap is the AIC~\citep{akaike1974new}, which is simply the number of degrees of freedom divided by the number of samples: $\displaystyle\hat{\gG}(\theta) = \tfrac{1}{N} d$ where $d$ is the dimensionality of $\theta$. This estimator is valid locally around the maximum likelihood parameters computed on the training data.  However, these assumptions do not hold in most cases, leading to the number of parameters being a poor predictor of the generalization gap~\citep{novak2018sensitivity}.
When dealing with maximum likelihood estimation (MLE) in misspecified models, a more general formula for estimating the gap is given by the Takeuchi information criterion~\citep[TIC:][]{takeuchi1976distribution}:
\begin{align}
\label{eq:tic}
  \hat{\gG} &= \frac{1}{N} \Tr(\rmH(\hat\theta^\ast)^{-1} \rmC(\hat\theta^\ast))  \; ,
\end{align}
where $\hat\theta^\ast$ is a local optimum. Note that $\rmH$ and $\rmC$ here are the hessian and covariance of the gradients matrices computed on the \emph{true data distribution}.

This criterion is not new in the domain of machine learning. It was rediscovered by~\citet{murata1994network} and similar criteria have been proposed since then~\citep{beirami2017optimal, wang2018approximate}. However, as far as we know, no experimental validation of this criterion has been carried out for deep networks.
Indeed, for deep networks, $\rmH$ is highly degenerate, most of its eigenvalues being close to $0$~\citep{sagun2016eigenvalues}. In this work, the Takeuchi information criterion is computed, in the degenerate case, by only taking into account the eigenvalues of the Hessian of significant magnitude. In practice, we cut all the eigenvalues smaller than a constant times the biggest eigenvalue and perform the inversion on that subspace. Details can be found in appendix~\ref{app:inversion}.

Interestingly, the term $\Tr(\rmH^{-1}\rmC)$ appeared in several works before, whether as an upper bound on the suboptimality~\citep{flammarion2015averaging} or as a number of iterates required to reach a certain suboptimality~\citep{bottou2008tradeoffs}. Sadly, it is hard to estimate for large networks but we propose an efficient approximation in Section~\ref{sec:tic_approx}.

\subsection{Limitations of flatness and sensitivity}
We highlight here two commonly used estimators of the generalization gap as they provide good examples of failure modes that can occur when not taking the noise into account. This is not to mean that these estimators cannot be useful for the models that are common nowadays, rather that they are bound to fail in some cases.

\textbf{Flatness}~\citep{hochreiter1997flat} links the spectrum of the Hessian at a local optimum with the generalization gap. This correlation, observed again by~\citet{keskar2016large}, was already shown to not hold in general~\citep{dinh2017sharp}. As we showed in Section~\ref{sec:C_is_not_F}, the Hessian does not capture the covariance of the data, which is linked to the generalization gap through the central-limit theorem.

\textbf{Sensitivity}~\citep{novak2018sensitivity} links the generalization gap to the derivative of the loss with respect to the input. The underlying idea is that we can expect some discrepancy between train and test data, which will induce changes in the output and a potentially higher test loss. However, penalizing the norm of the Jacobian assumes that changes between train and test data will be isotropic. In practice, we can expect data to vary more along some directions, which is not reflected in the sensitivity. In the extreme case where the test data is exactly the same as the training data, the generalization gap will be 0, which will again not be captured by the sensitivity. In practice, whitening the data makes the sensitivity appropriate, save for a scaling factor, as we will see in the experiments.

\section{Experiments}
We now provide experimental validation of all the results in this paper. We start by analyzing the distance between information matrices, first showing its poor correlation with the training loss, then showing that these matrices appear to be remarkably aligned, albeit with different scales, when training deep networks on standard datasets.

\subsection{Discrepancies between \texorpdfstring{$\rmC$}{C}, \texorpdfstring{$\rmH$}{H} and \texorpdfstring{$\rmF$}{F}}
\label{sec:exp_hfc}

\subsubsection{Experimental setup}
\label{sec:exp_setup_hfc}
For comparing the similarities and discrepencies between the information matrices, we tested
\begin{itemize}[leftmargin=*]
    \item 5 different architectures: logistic regression, a 1-hidden layer and 2-hidden layer fully connected network, and 2 small convolutional neural networks (CNNs, one with batch normalization~\citep{ioffe2015batch} and one without);
    \item 3 datasets: MNIST, CIFAR-10, SVHN;
    \item 3 learning rates: $10^{-2}$, $5\cdot 10^{-3}$, and $10^{-3}$, using SGD with momentum $\mu = 0.9$;
    \item 2 batch sizes: $64$, $512$;
    \item 5 dataset sizes: 5k, 10k, 20k, 25k, and 50k.
\end{itemize}
We train for 750k steps and compute the metrics every 75k steps. To be able to compute all the information matrices exacly, we reduced the input dimension by converting all images to greyscale and resizing them to $7 \times 7$ pixels. While this makes the classification task more challenging, our neural networks still exhibit the behaviour of larger ones by their ability to fit the training set, even with random labels. Details and additional figures can be found in appendix~\ref{app:exp_details}.

\subsection{Comparing Fisher and empirical Fisher}

\begin{figure}[ht]
  \centering
    \includegraphics[width=.5\textwidth]{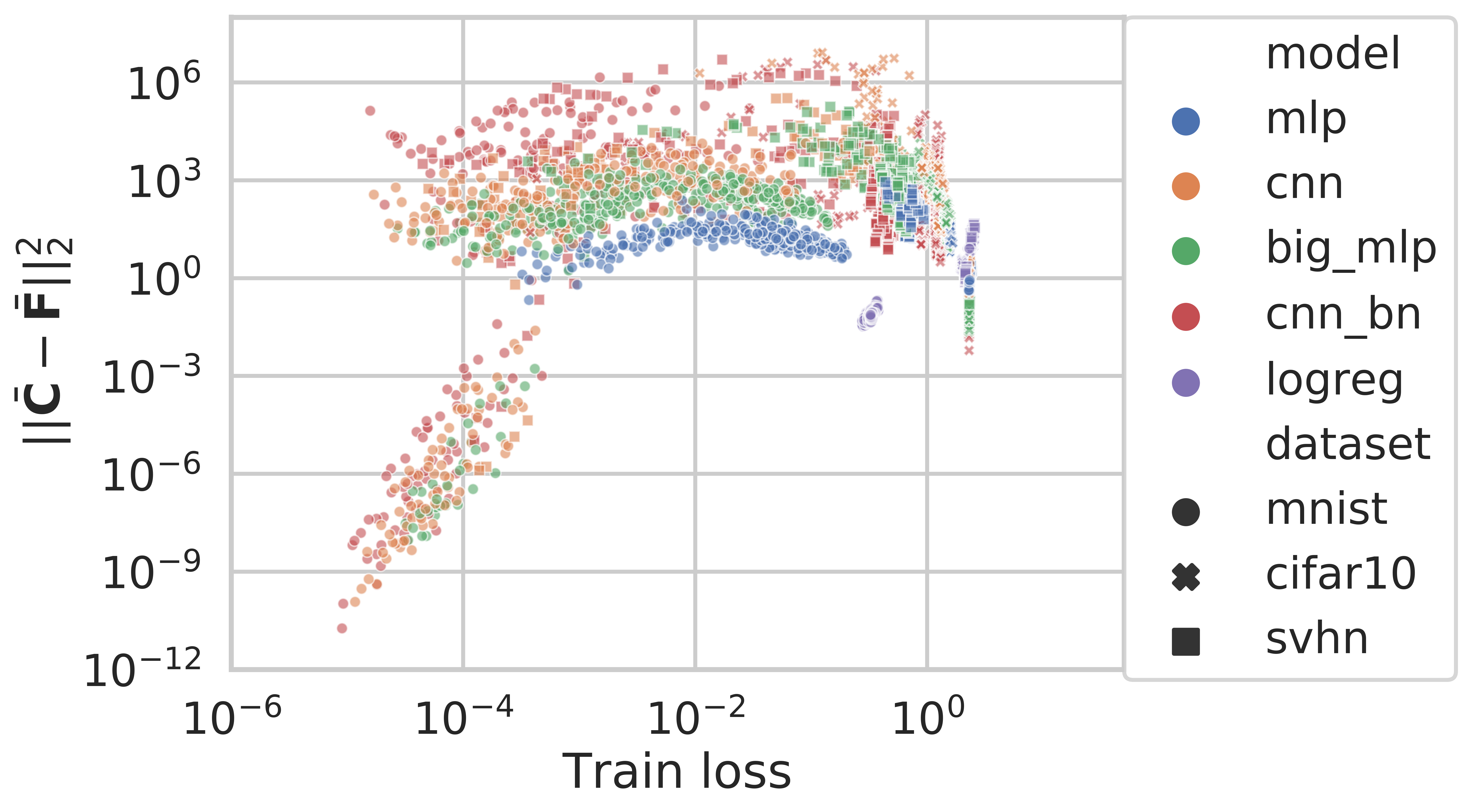}
    \caption{Squared Frobenius norm between $\bar{\rmF}$ and $\bar{\rmC}$ (computed on the training distribution). Even for some low training losses, there can be a significant difference between the two matrices.
    \label{fig:F_vs_C}}
\end{figure}

Figure~\ref{fig:F_vs_C} shows the squared Frobenius norm between $\rmF$ and $\rmC$ (on training data) for many architectures, datasets, at various stages of the optimization. We see that, while the two matrices eventually coincide on the training set for some models, the convergence is very weak as even low training errors can lead to a large discrepancy between these two matrices. In practice, $\rmC$ and $\rmF$ might be significantly different, even when computed on the training set.

\subsection{Comparing \texorpdfstring{$\rmH$}{H}, \texorpdfstring{$\rmF$}{F} and \texorpdfstring{$\rmC$}{C}}

\label{sec:HFC_diff}

In this subsection, we analyze the similarities and differences between the information matrices. We will focus on the scale similarity $r$, defined as the ratio of traces, and the angle similarity $s$, defined as the cosine between matrices.
Note that having both $r(\rmA, \rmB) = 1$ and $s(\rmA, \rmB) = 1$ implies $\rmA = \rmB$.

\begin{figure}[ht]

    \includegraphics[width=.23\textwidth]{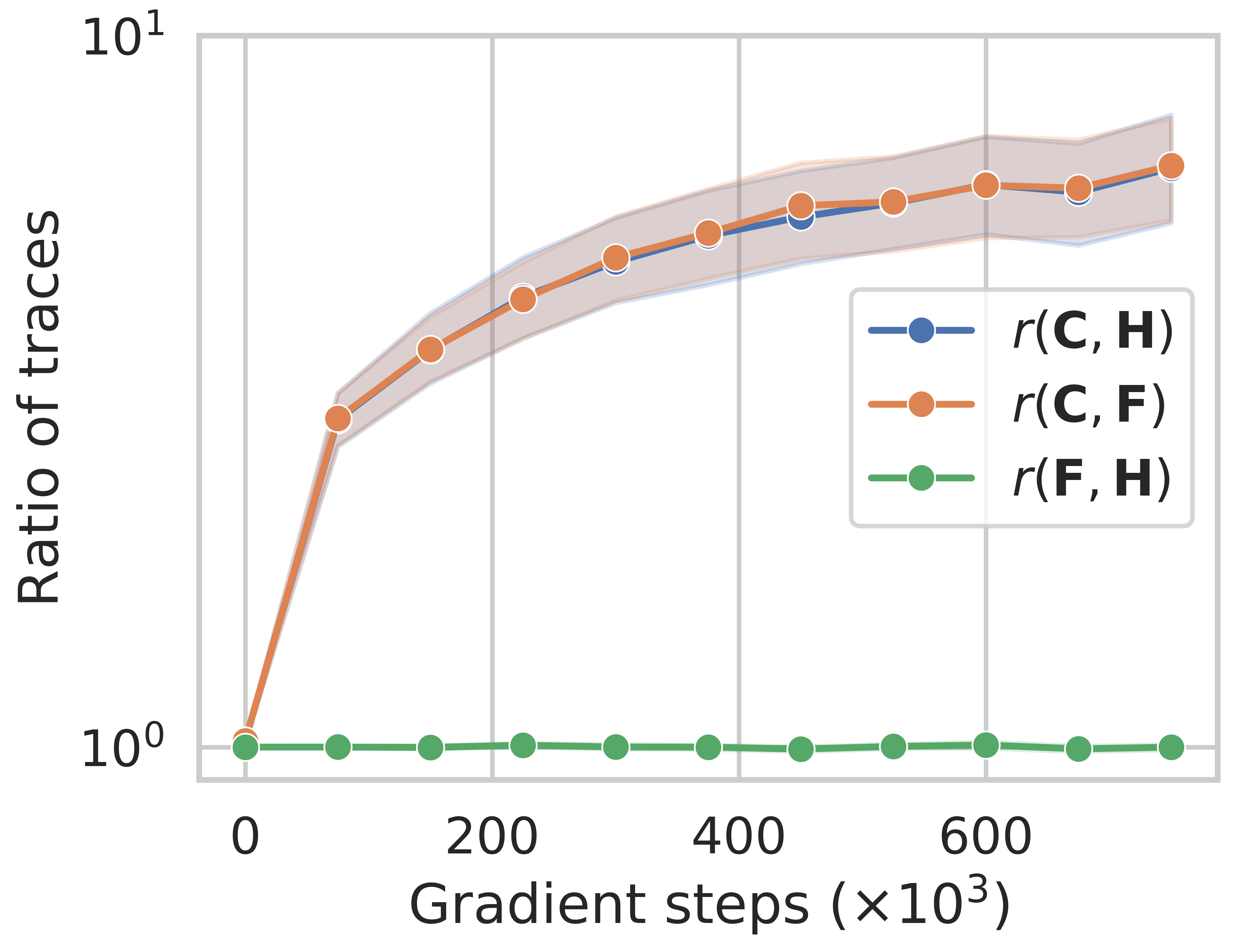}
      \hfill
    \includegraphics[width=.23\textwidth]{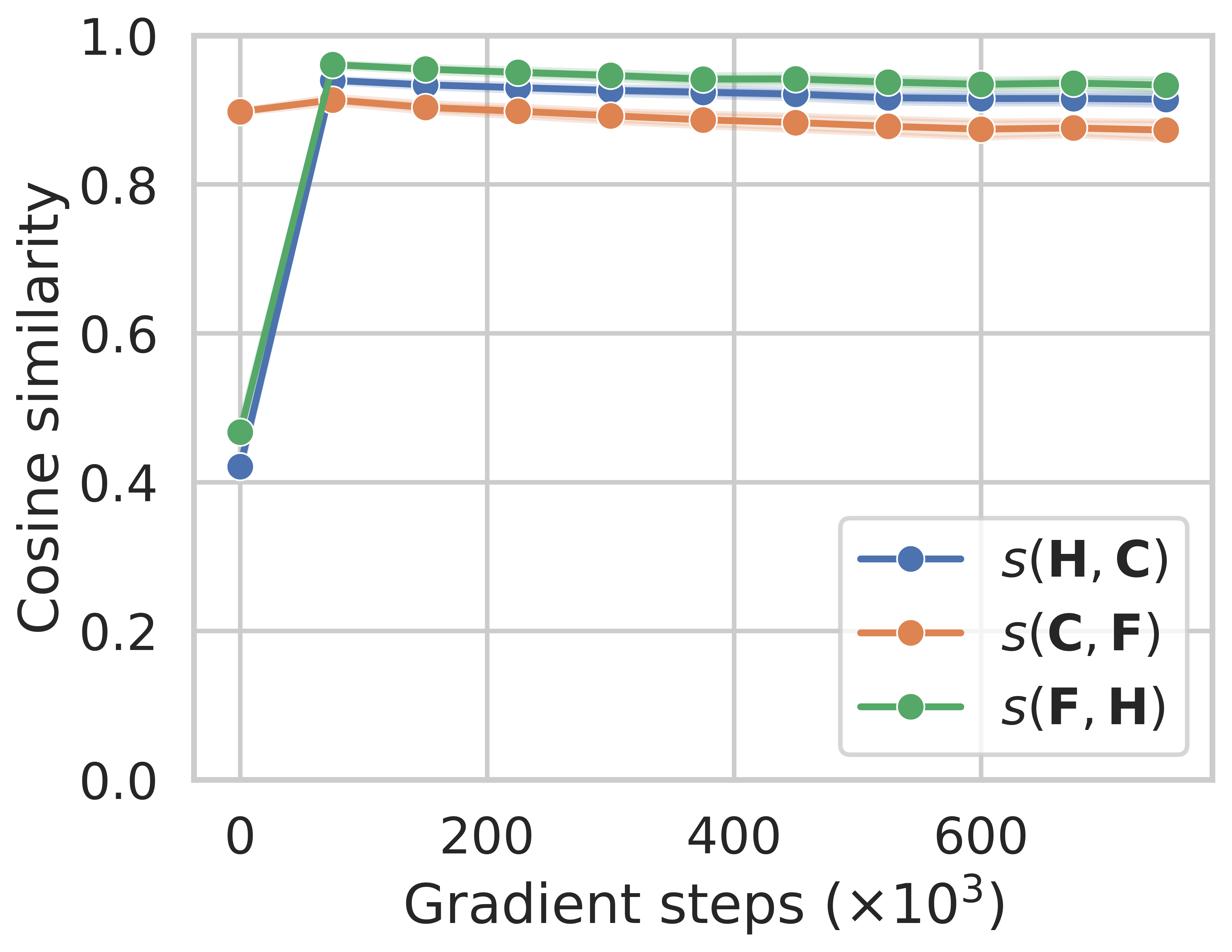}
    \caption{Scale and angle similarities between information matrices.
        \label{fig:similarities}}
\end{figure}

Figure~\ref{fig:similarities} shows the scale (left) and angle (right) similarities between the three pairs of matrices during the optimization of all models used in figure~\ref{fig:large_scale_tic}. We can see that $\rmH$ is not aligned with $\rmC$ nor $\rmF$ at the beginning of the optimization but this changes quickly. Then, all three matrices reach a very high cosine similarity, much higher than we would obtain for two random low-rank matrices. For the scaling, $\rmC$ is ``larger'' than the other two while $\rmF$ and $\rmH$ are very close to each other. Thus, as in the least squares case, we have $\rmC \appropto \rmF \approx \rmH$.

\subsection{Impact of noise on second-order methods}
Section~\ref{sec:optimization} extended existing results to take the geometry of the noise and the curvature into account. Here, we show how the geometry of the noise, and in particular its relationship to the Hessian, can make or break second-order methods in the stochastic setting. To be clear, we assume here that we have access to the full Hessian and do not address the issue of estimating it from noisy samples.

We assume a quadratic $\ell(\theta) = \tfrac{1}{2} \theta^\top \rmH \theta$ with $\theta \in \R^{20}$ and $\rmH \in \R^{20\times 20}$ a diagonal matrix such that $\rmH_{ii} = i^2$ with a condition number $d^2 = 400$. At each timestep, we have access to an oracle that outputs a noisy gradient, $\rmH\theta_t + \epsilon$ with $\epsilon$ drawn from a zero-mean Gaussian with covariance $\rmS$. Note here that $\rmS$ is the \emph{centered} covariance of the gradients. We consider three settings: a) $\rmS = \alpha_1 \rmH$; b) $\rmS = \rmI$; c) $\rmS = \alpha_{-1} \rmH^{-1}$ where the constants $\alpha_1$ and $\alpha_{-1}$ are chosen such that $\Tr(\rmS) = d$. Hence, these three settings are indistinguishable from the point of view of the rate of~\citet{schmidt2014convergence}.

In this simplified setting, we get an analytic formula for the variance at each timestep and we can compute the exact number of steps $t$ such that $\E[\Delta_t]$ falls below a suboptimality threshold. To vary the impact of the noise, we compute the number of steps for three different thresholds: a) $\epsilon = 1$; b) $\epsilon = 0.1$; c) $\epsilon = 0.01$. For each algorithm and each noise, the stepsize is optimized to minimize the number of steps required to reach the threshold.

The results are in Table~\ref{tab:iter_noise}. We see that, while Stochastic gradient and momentum are insensitive to the geometry of the noise for small $\epsilon$, Newton method is not and degrades when the noise is large in low curvature directions. For $\epsilon = 10^{-2}$ and $\rmS \propto \rmH^{-1}$, Newton is worse than SG, a phenomenon that is not captured by the bounds of~\citet{bottou2008tradeoffs} since they do not take the structure of the curvature and the noise into account. We also see that the advantage of Polyak momentum over stochastic gradient disappears when the suboptimality is small, i.e. when the noise is large compared to the signal.

Also worthy of notice is the fixed stepsize required to achieve suboptimality $\epsilon$, as shown in Table~\ref{tab:stepsize_noise}. While it hardly depends on the geometry of the noise for SG and Polyak, Newton method requires much smaller stepsizes when $\rmS$ is anticorrelated with $\rmH$ to avoid amplifying the noise.

\begin{table}
\centering
\begin{tabular}{|c|c||c|c|c|}
\hline
$\epsilon$&\textbf{Method} &   $\beta=1$ &   $\beta=0$ &   $\beta=-1$\\
\hline
\multirow{3}{*}{$10^0$}&SG              &   44 &   43 &   42\\
&Newton                                 &   3  &   2   &   19\\
&Polyak                                 &   36  &   36  &   34\\
\hline
\multirow{3}{*}{$10^{-1}$}&SG   &   288     &   253 &   207\\
&Newton                         &   3       &   28  &   225\\
&Polyak                         &   119     &   111 &   97\\
\hline
\multirow{3}{*}{$10^{-2}$}&SG   &   2090    &   1941    &   1731\\
&Newton                         &   29      &   315     &   2663\\
&Polyak                         &   1743    &   1727    &   1705\\
\hline
\end{tabular}
\caption{Number of updates required to reach suboptimality of $\epsilon$ for various methods and $\rmS \propto \rmH^\beta$.
\label{tab:iter_noise}}
\end{table}

\begin{table}
\centering
\begin{tabular}{|c|c||c|c|c|}
\hline
$\epsilon$&\textbf{Method} &   $\beta=1$ &   $\beta=0$ &   $\beta=-1$\\
\hline
\multirow{3}{*}{$10^0$}&SG              &   $5\cdot 10^{-3}$ &   $5\cdot 10^{-3}$ &   $5\cdot 10^{-3}$\\
&Newton                                 &   $1\cdot 10^{0}$  &   $1\cdot 10^0$   &   $2\cdot 10^{-1}$\\
&Polyak                                 &   $5\cdot 10^{-3}$  &   $4\cdot 10^{-3}$  &   $5\cdot 10^{-3}$\\
\hline
\multirow{3}{*}{$10^{-1}$}&SG   &   $4\cdot 10^{-3}$     &   $4\cdot 10^{-3}$ &   $5\cdot 10^{-3}$\\
&Newton                         &   $1\cdot 10^0$       &   $2\cdot 10^0$  & $3\cdot 10^{-2}$\\
&Polyak                         &   $2\cdot 10^{-3}$     &   $2\cdot 10^{-3}$ &   $3\cdot 10^{-3}$\\
\hline
\multirow{3}{*}{$10^{-2}$}&SG   &   $1\cdot 10^{-3}$    &   $1\cdot 10^{-3}$    &   $2\cdot 10^{-3}$\\
&Newton                         &   $2\cdot 10^{-1}$      &   $2\cdot 10^{-2}$     &   $3\cdot 10^{-3}$\\
&Polyak                         &   $3\cdot 10^{-4}$    &   $3\cdot 10^{-4}$    &   $3\cdot 10^{-4}$\\
\hline
\end{tabular}
\caption{Stepsizes achieving suboptimality $\epsilon$ in the fewest updates for various methods and $\rmS \propto \rmH^\beta$.
\label{tab:stepsize_noise}}
\end{table}

\subsection{The TIC and the generalization gap}
\label{sec:neyshabur_gen}
We now empirically test the quality of the TIC as an estimator of the generalization gap in deep networks.
Following~\citet{neyshabur2017exploring} we assess the behaviour of our generalization gap estimator by varying (1) the number of parameters in a model and (2) the label randomization ratio.

\begin{figure}[ht]
 \centering
  \begin{subfigure}[b]{.46\textwidth}
    \includegraphics[trim={0 0 0 2mm},clip, width=\textwidth]{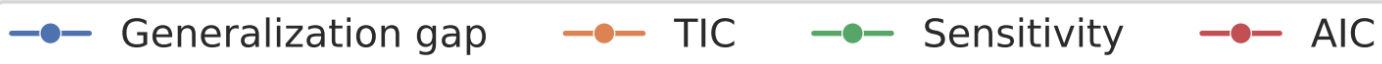}
  \end{subfigure}
  \centering
  \begin{subfigure}[b]{0.23\textwidth}
    \includegraphics[trim={0 0 8mm 0},clip,width=\textwidth]{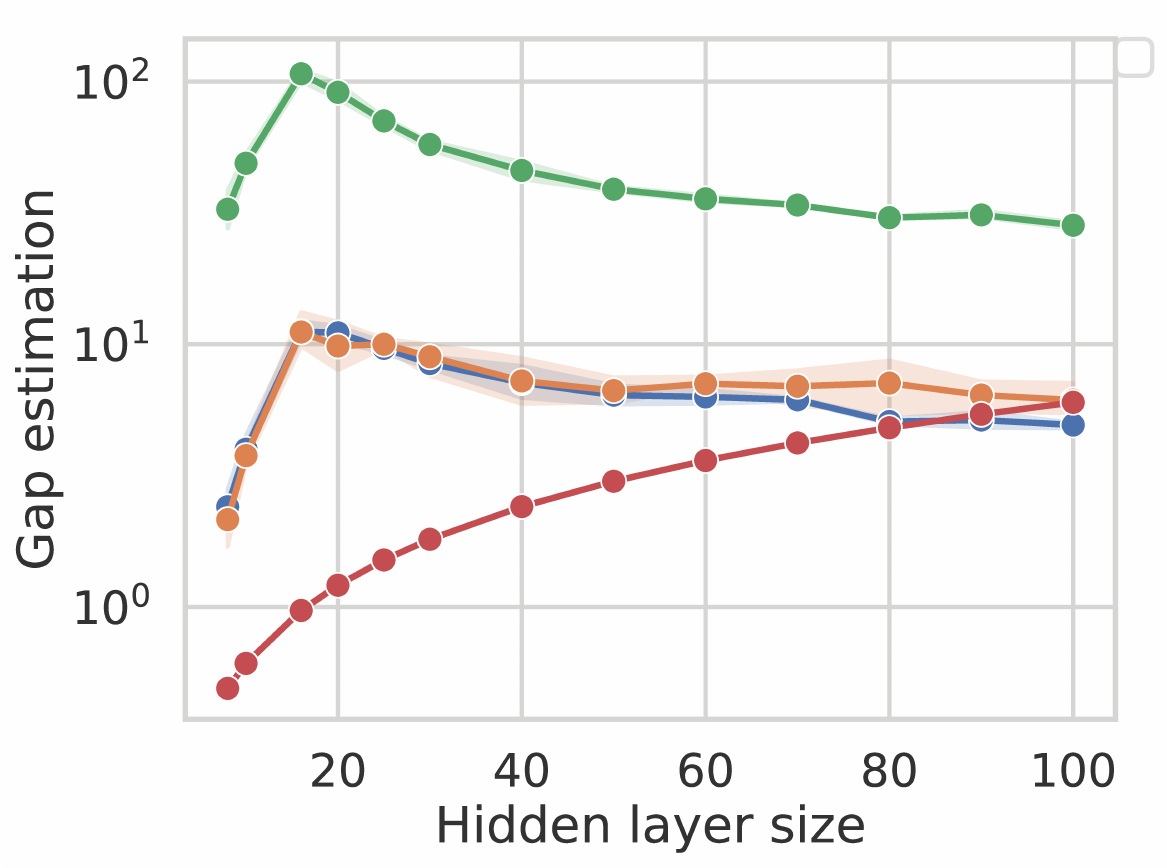}
    \caption{Varying hidden layer size.}
    \label{fig:vary_h}
  \end{subfigure}
  \hfill
  \begin{subfigure}[b]{0.23\textwidth}
    \includegraphics[trim={0 0 8mm 0},clip,width=\textwidth]{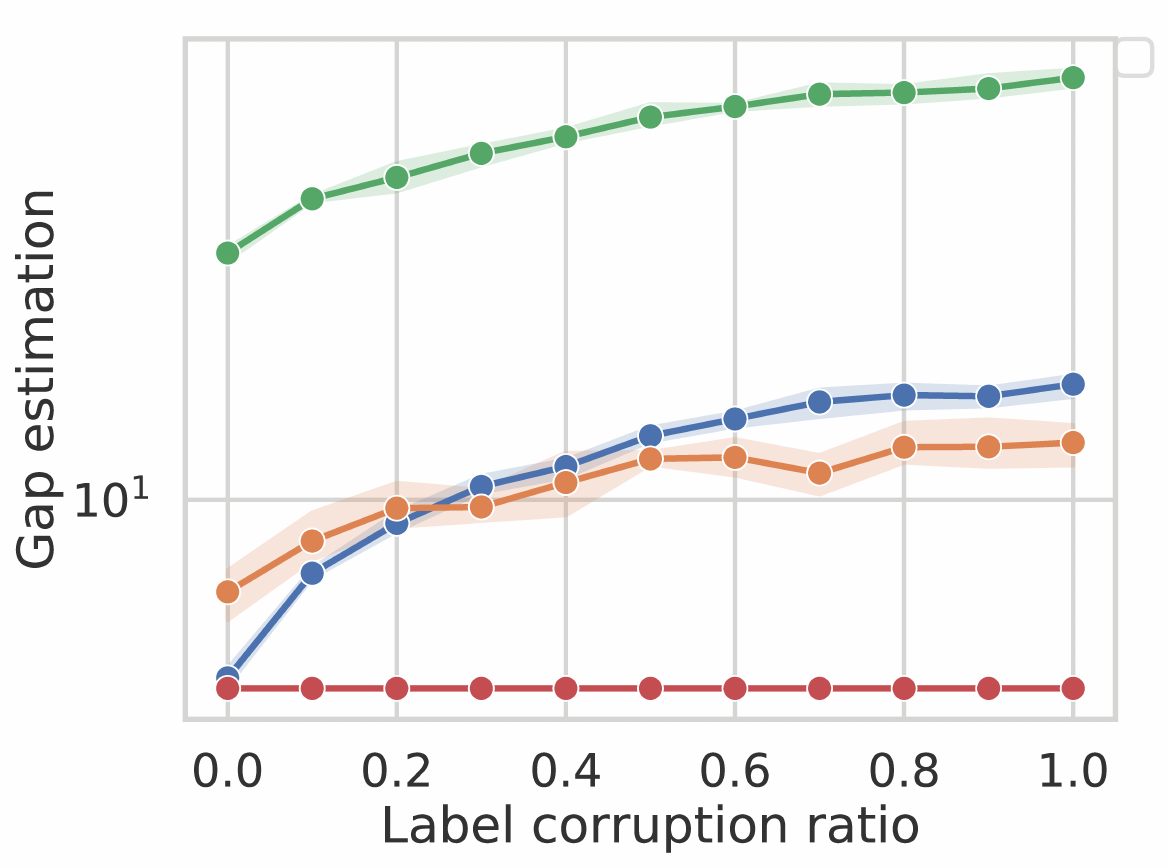}
    \caption{Varying the label randomization level.}
    \label{fig:vary_corr}
  \end{subfigure}
 
  \label{fig:neyshabur_exp}
\caption{Comparing the TIC to other estimators of the generalization gap on SVHN. The TIC matches the generalization gap more closely than both the AIC and the sensitivity.}
\end{figure}

Experiments are performed using a fully connected feedforward network with a single hidden layer trained on a subset of $2$k samples of SVHN~\citep{netzer2011reading}. In Figure~\ref{fig:vary_h} we vary the number of units in the hidden layer without label randomization while in Figure~\ref{fig:vary_corr} we vary the label randomization ratio with a fixed architecture.
Each point is computed using 3 different random number generator seeds. The neural networks are trained for 750k steps. The confidence intervals are provided using bootstrapping to estimate a $95 \%$ confidence interval. The Hessian, covariance matrices and sensitivity are computed on a subset of size $5$k of the test data. Details can be found in Appendix~\ref{app:exp_details}.

\begin{figure}[ht]
\centering
\begin{subfigure}[b]{.46\textwidth}
    \includegraphics[trim={0mm 0mm 0mm 0mm},clip,width=\textwidth]{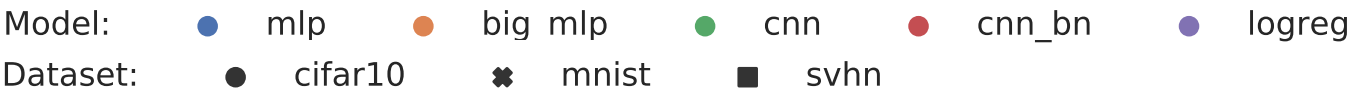}
  \end{subfigure}
  \centering
  \begin{subfigure}[b]{0.23\textwidth}
    \includegraphics[trim={0 0 60mm 0},clip,width=\textwidth]{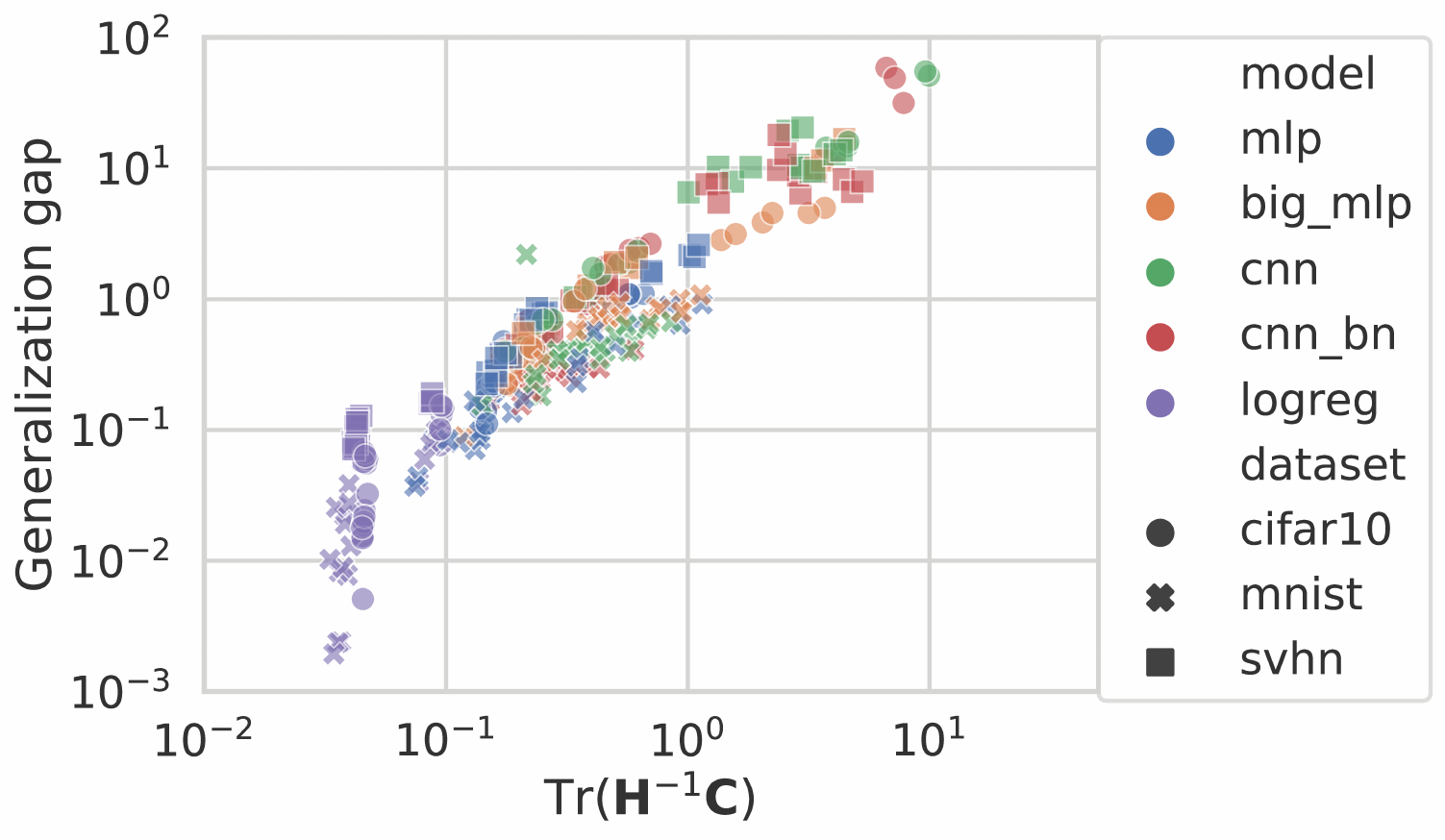}
    \caption{ Gap vs. TIC.
    \label{fig:tic_te}}
  \end{subfigure}
    \hfill
      \begin{subfigure}[b]{0.23\textwidth}
    \includegraphics[trim={0 0 55mm 0},clip,width=\textwidth]{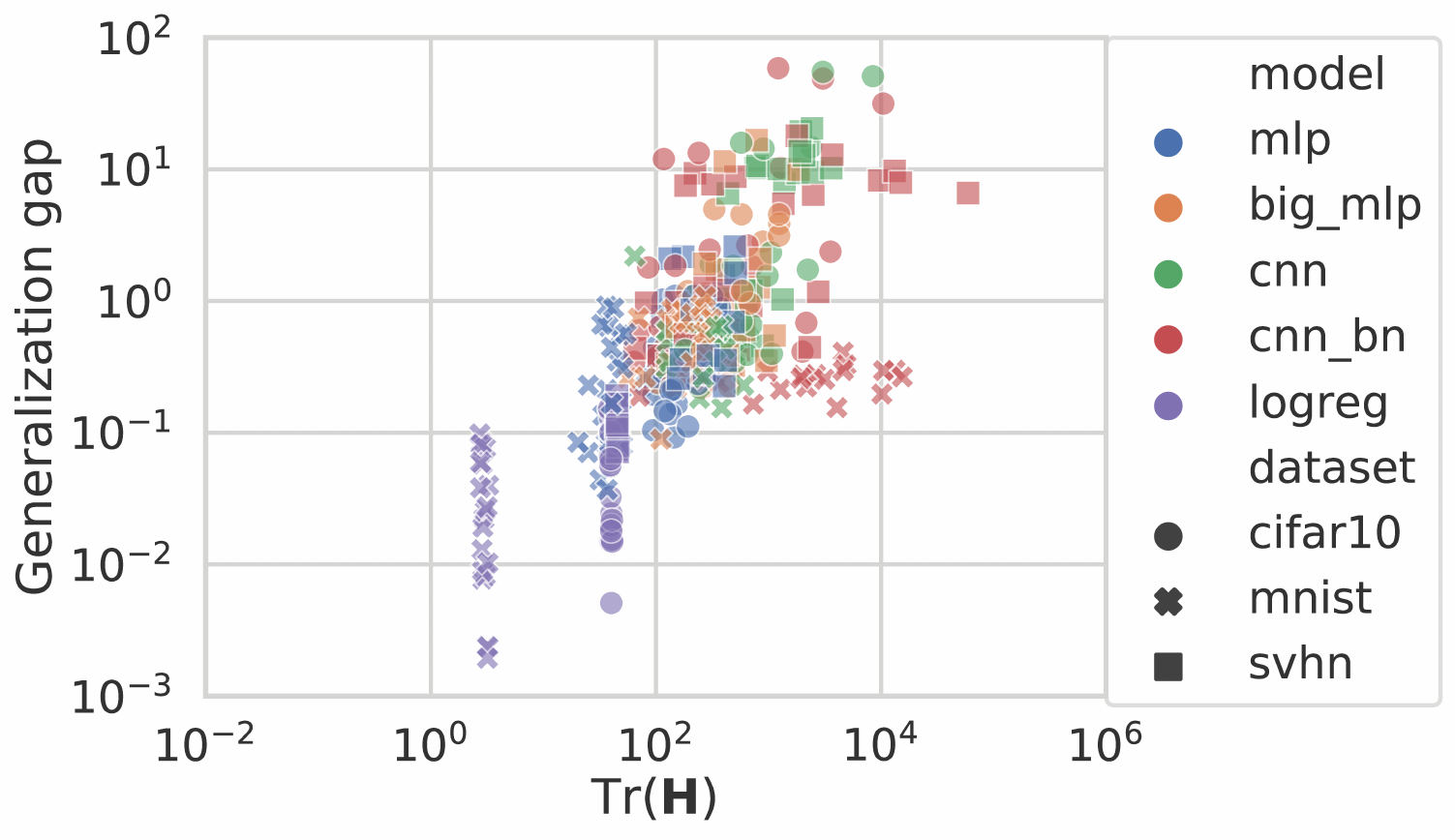}
    \caption{ Gap vs. flatness.
    \label{fig:flatness}}
  \end{subfigure}
  
      \caption{Generalization gap as a function of the Takeuchi information criterion (\emph{left}) and the trace of the Hessian on the test set (\emph{right}) for many architectures, datasets, and hyperparameters. Correlation is perfect if all points lie on a line. We see that the Hessian cannot by itself capture the generalization gap.}
      \label{fig:large_scale_tic}
\end{figure}

We now study the ability of the TIC across a wide variety of models, datasets, and hyperparameters. More specifically, we compare the TIC to the generalization gap for:
The experiments of Figure~\ref{fig:large_scale_tic} are performed with the experimental setup presented in subsection~\ref{sec:exp_setup_hfc}.
Figure~\ref{fig:tic_te} shows that the TIC using $\rmH$ and $\rmC$ computed over the test set is an excellent estimator of the generalization gap. For comparison, we also show in Figure~\ref{fig:flatness} the generalization gap as a function of $\rmH$ computed over the test set. We see that, even when using the test set, the correlation is much weaker than with the TIC.

\subsubsection{Efficient approximations to the TIC}
\label{sec:tic_approx}
Although the TIC is a good estimate of the generalization gap, it can be expensive to compute on large models. Following our theoretical and empirical analysis of the proximity of $\rmH$ and $\rmF$, we propose two approximations to the TIC: $\Tr(\rmF^{-1} \rmC)$ and $\Tr(\rmC) / \Tr(\rmF)$. They are easier to compute as the $\rmF$ is in general easier to compute than $\rmH$ and the second does not require any matrix inversion.

\begin{figure}[ht]
\centering
    \includegraphics[trim={00mm 0mm 0mm 0mm},clip,width=.4\textwidth]{./figures/legend_gimp.png}
    \includegraphics[trim={0 0 60mm 0},clip,width=.23\textwidth]{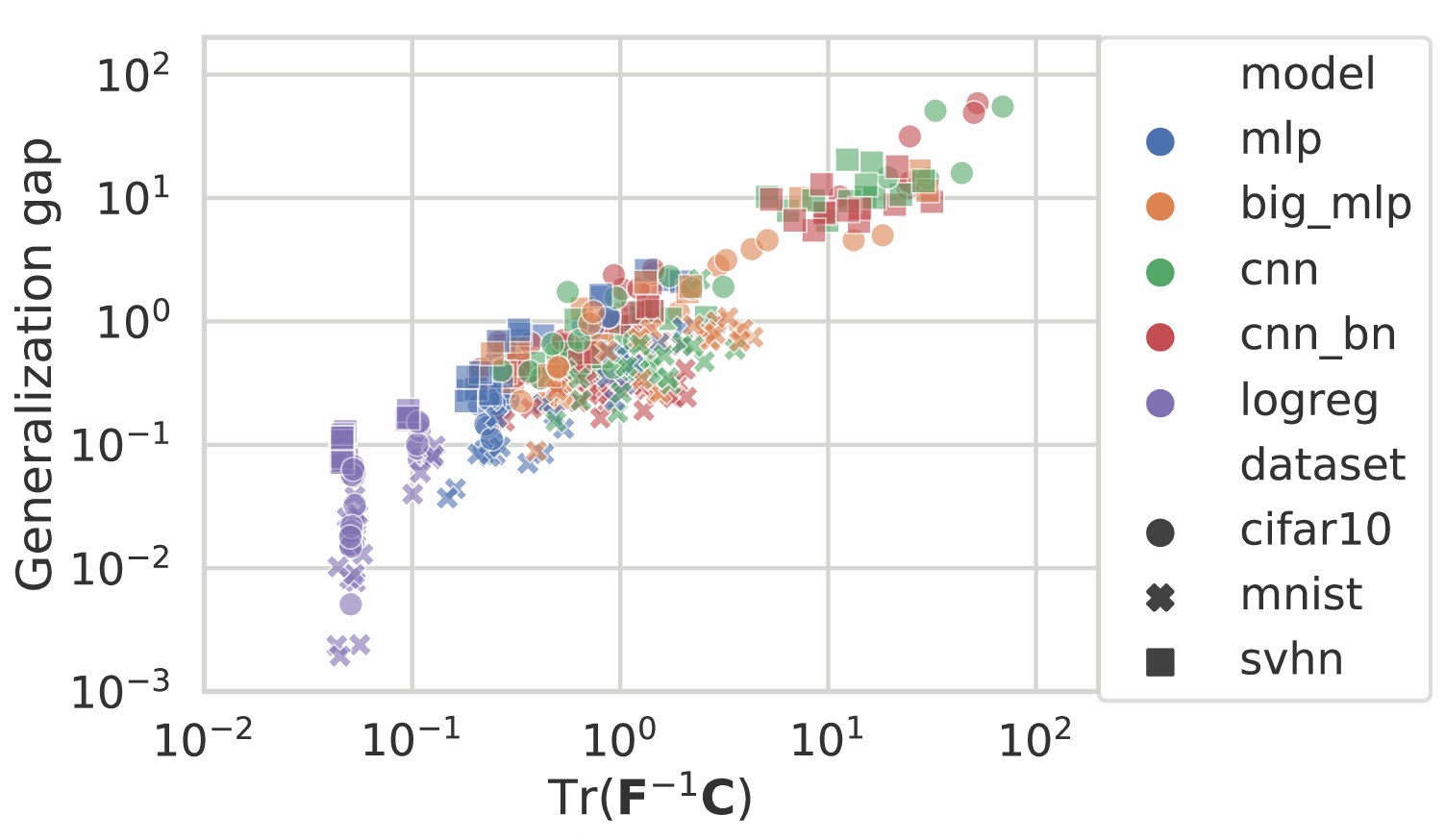}
    \includegraphics[trim={0 0 60mm 0},clip,width=.23\textwidth]{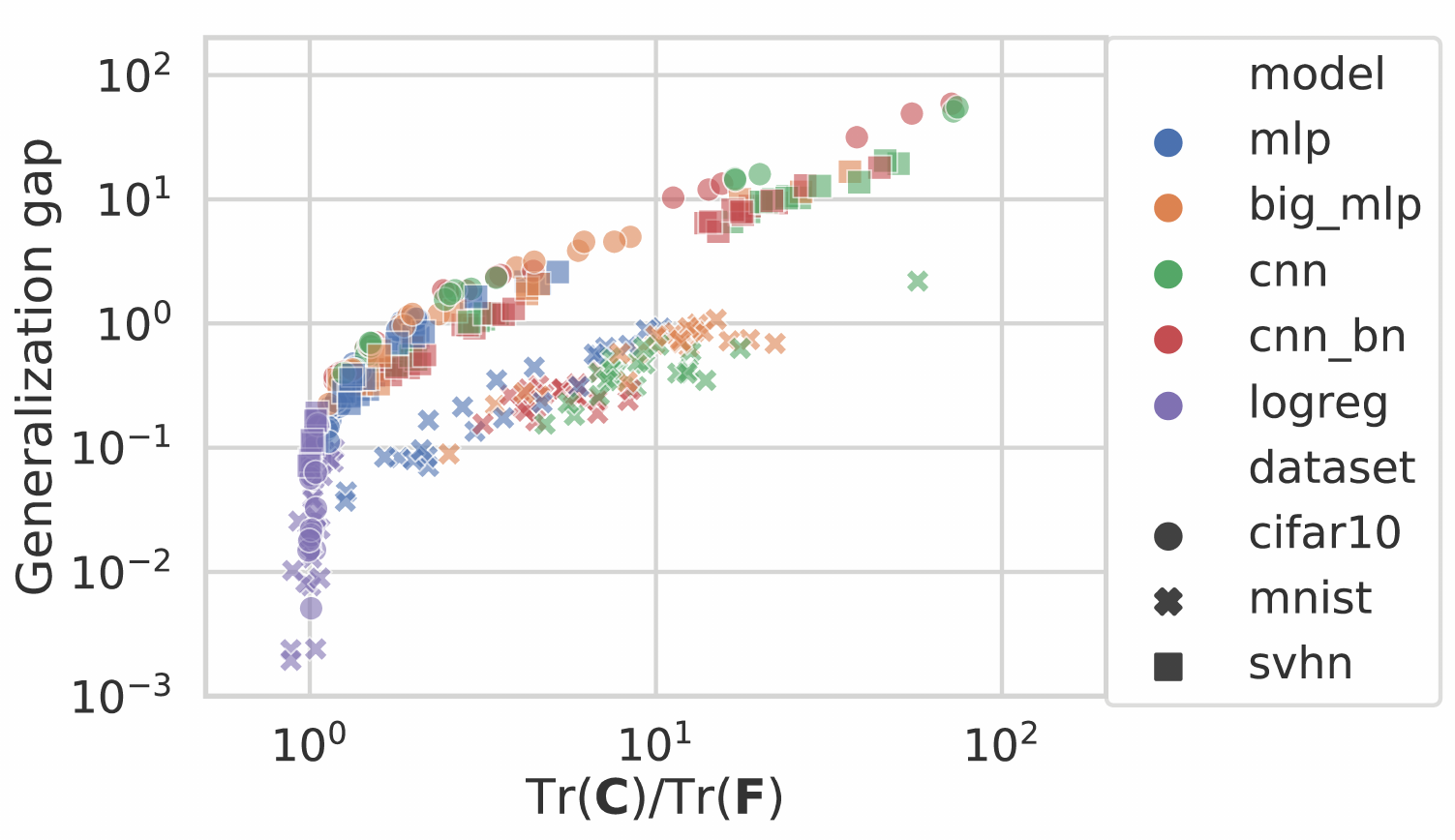}
\caption{Generalization gap as a function of two approximations to the Takeuchi Information Criterion: $\Tr(\rmF^{-1} \rmC)$ (\emph{left}) and $\Tr(\rmC) / \Tr(\rmF)$ (\emph{right}).
\label{fig:tic_approx}}
\end{figure}

Using the same experimental setting as in~\ref{sec:neyshabur_gen}, we observe in Figure~\ref{fig:tic_approx} that the replacing $\rmH$ with $\rmF$ leads to almost no loss in predictive performance. On the other hand, the ratio of the traces works best when the generalization gap is high and tends to overestimate it when it is small.

\textbf{Intuition on $\Tr(\rmC) / \Tr(\rmF)$}: it is not clear right away why the ratio of traces might be a interesting quantity. However, as observed in figure~\ref{fig:similarities}, $\rmC$ and $\rmF$ are remarkably aligned, but there remains a scaling factor. If we had $\rmC = \alpha \rmF$, then $\Tr(\rmF^{-1} \rmC) = k \alpha$ where $k$ is the dimension of the invertible subspace of $\rmF$ and $\Tr(\rmC) / \Tr(\rmF) = d \alpha$ where $d$ is the dimensionality of $\theta$. So, up to a multiplicative constant (or an offset in log scale), we can expect these two quantities to exhibit similarities.
Notice that on figure~\ref{fig:tic_approx}, this offset does appear and is different for every dataset (MNIST has the smallest one, then SVHN and CIFAR10, just slightly bigger).

\subsection{The importance of the noise in estimating the generalization gap}
For a given model, the generalization gap captures the discrepancy that exists between the training set and the data distribution. Hence, estimating that gap involves the evaluation of the uncertainty around the data distribution. The TIC uses $\rmC$ to capture that uncertainty but other measures probably exist. However, estimators which do not estimate it are bound to have failure modes. For instance, by using the square norm of the derivative of the loss with respect to the input, the sensitivity implicitly assumes that the uncertainty around the inputs is isotropic and will fail should the data be heavily concentrated in a low-dimensional subspace. It would be interesting to adapt the sensitivity to take the covariance of the inputs into account.

Another aspect worth mentioning is that estimators such as the margin assume that the classifier is fixed but the data is a random variable. Then, the margin quantifies the probability that a new datapoint would fall on the other side of the decision boundary. By contrast, the TIC assumes that the data are fixed but that the classifier is a random variable. It estimates the probability that a classifier trained on slightly different data would classify a training point incorrectly. In that, it echoes the uniform stability theory~\citep{bousquet2002stability}, where a full training with a slightly different training set has been replaced with a local search.

\section{Conclusion and open questions}
We clarified the relationship between information matrices used in optimization. While their differences seem obvious in retrospect, the widespread confusion makes these messages necessary. Indeed, several well-known algorithms, such as Adam~\citep{kingma2014adam}, claiming to use second-order information about the loss to accelerate training seem instead to be using the covariance matrix of the gradients. Equipped with this new understanding of the difference between the curvature and noise information matrices, one might wonder if the success of these methods is not due to variance reduction instead. If so, one should be able to combine variance reduction and geometry adaptation, an idea attempted by~\citet{leroux2011}.

We also showed how, in certain settings, the geometry of the noise could affect the performance of second-order methods. While Polyak momentum is affected by the scale of the noise, its performance is independent of the geometry, similar to stochastic gradient but unlike Newton method. However, empirical results indicate that common loss functions are in the regime favorable to second-order methods.

Finally, we investigated whether the Takeuchi information criterion is relevant for estimating the generalization gap in neural networks. We provided evidence that this complexity measures involving the information matrices is predictive of the generalization performance.

We hope this study will clarify the interplay of the noise and curvature in common machine learning settings, potentially giving rise to new optimization algorithms as well as new methods to estimate the generalization gap.

\subsubsection*{Acknowledgments}
We would like to thank Gauthier Gidel, Reyhane Askari and Giancarlo Kerg for reviewing an earlier version of this paper. We also thank Aristide Baratin for insightful discussions.
Valentin Thomas acknowledges funding from the Open Philantropy project.

\bibliography{bibfile}
\bibliographystyle{apalike}
\clearpage

\appendix
\onecolumn

\section{Proofs}

\subsection{Bounds between \texorpdfstring{$\rmH$}{H}, \texorpdfstring{$\rmF$}{F} and \texorpdfstring{$\rmC$}{C}}
\label{app:bounds}

\subsubsection{Bounds with backward \texorpdfstring{$\chi^2$}{chi-2} divergence}
\begin{eqnarray*}
|\rmF_{ij} - \rmH_{ij}|^2 &=& |\int \q(x, y) \big(\nabla_\theta^2 \ell(x, y)\big)_{ij} d(x, y) - \int p(x, y) \big(\nabla_\theta^2 \ell(x, y)\big)_{ij} d(x, y)|^2\\
&=& |\int \big(\q(x, y) - p(x, y)\big) \big(\nabla_\theta^2 \ell(x, y)\big)_{ij} d(x, y)|^2\\
&=& |\int \frac{\big(\q(x, y) - p(x, y)\big)}{\sqrt{p(x, y)}} \big(\sqrt{p(x, y)} \nabla_\theta^2 \ell(x, y)\big)_{ij} d(x, y)|^2\\
&\le& \int \frac{\big(\q(x, y) - p(x, y)\big)^2}{p(x, y)} d(x, y) \int p(x, y)  \big(\nabla_\theta^2 \ell(x, y)\big)_{ij}^2 d(x, y)\\
&=& \gD_{\chi^2}(\q||p) \ \E_p [ \big(\nabla_\theta^2 \ell(x, y)\big)_{ij}^2 ]
\end{eqnarray*}
Where we used Cauchy-Schwarz inequality and $\gD_{\chi^2}$ denotes the $\chi^2$ divergence.

\begin{eqnarray*}
||\rmF - \rmH||^2 &\le& \gD_{\chi^2}(\q||p) \ \E_p [ ||\rmH(x, y)||^2_2 ]
\end{eqnarray*}
Where $\rmH(x, y) \triangleq \nabla_\theta^2 \ell(x, y)$ is the empirical hessian for one sample and the $|| \cdot ||_2$ is the Frobenius norm.

In the same way
\begin{eqnarray*}
|\rmF_{ij} - \rmC_{ij}|^2 &=& |\int \q(x, y) \big(\nabla_\theta \ell(x, y) \nabla_\theta \ell(x, y)^\top \big)_{ij} d(x, y) - \int p(x, y) \big(\nabla_\theta \ell(x, y) \nabla_\theta \ell(x, y)^\top \big)_{ij} d(x, y)|^2\\
&\le& \gD_{\chi^2}(\q||p) \ \E_p [ \big(\nabla_\theta \ell(x, y) \nabla_\theta \ell(x, y)^\top \big)_{ij}^2 ]
\end{eqnarray*}

For $\rmC(x, y) \triangleq \nabla_\theta \ell(x, y) \nabla_\theta \ell(x, y)^\top$ we have

\begin{eqnarray*}
||\rmF - \rmC||^2 &\le& \gD_{\chi^2}(\q||p)\ \E_p [ ||\rmC(x, y)||^2 ]
\end{eqnarray*}

Hence
\begin{eqnarray*}
||\rmC - \rmH||^2 &\le& \gD_{\chi^2}(\q||p) \ \E_p [ ||\rmC(x, y)||^2  + ||\rmH(x, y)||^2]
\end{eqnarray*}

\subsubsection{Bounds with forward \texorpdfstring{$\chi^2$}{chi-2} divergence}
Note that in the above proof, breaking the integral in two with Cauchy-Schwarz inequality could have been done using

\begin{eqnarray*}
|\rmF_{ij} - \rmH_{ij}|^2 &=& |\int \frac{\big(\q(x, y) - p(x, y)\big)}{\sqrt{\q(x, y)}} \big(\sqrt{\q(x, y)} \nabla_\theta^2 \ell(x, y)\big)_{ij} d(x, y)|^2\\
&\le& \int \frac{\big(\q(x, y) - p(x, y)\big)^2}{\q(x, y)} d(x, y) \int \q(x, y)  \big(\nabla_\theta^2 \ell(x, y)\big)_{ij}^2 d(x, y)\\
&=& \gD_{\chi^2}(p||\q) \ \E_{\q} [ \big(\nabla_\theta^2 \ell(x, y)\big)_{ij}^2 ]
\end{eqnarray*}

Similarly
\begin{eqnarray*}
    |\rmF_{ij} - \rmC_{ij}|^2  &\le& \gD_{\chi^2}(p||\q) \ \E_{\q} [ \big(\nabla_\theta \ell(x, y) \nabla_\theta \ell(x, y)^\top \big)_{ij}^2 ]
\end{eqnarray*}
Thus 
\begin{eqnarray*}
||\rmC - \rmH||^2 &\le& \gD_{\chi^2}(p||\q) \ \E_{\q} [ ||\rmC(x, y)||^2  + ||\rmH(x, y)||^2]
\end{eqnarray*}

\subsubsection{Proof of Proposition \ref{prop:function_value}}
\label{proof:function_value}

From the upper bound assumption we have

\begin{eqnarray*}
f(\theta^{k+1}) &\le& f(\theta^{k}) + \nabla f(\theta^k)^\top (\theta^{k+1} - \theta^k) + \tfrac{1}{2} (\theta^{k+1} - \theta^k)^\top \rmH (\theta^{k+1} - \theta^k)\\
&=& f(\theta^{k}) - \alpha \nabla f(\theta^{k})^\top \rmM \nabla \ell(\theta^k, x) + \tfrac{\alpha^2}{2} \nabla \ell(\theta^k, x)^\top \rmM^\top \rmH \rmM \nabla \ell(\theta^k, x)\,. \\
\end{eqnarray*}

Subtracting $f(\theta^\ast)$ from both sides and taking conditional expectation we have
\begin{eqnarray*}
\E [ f(\theta^{k+1}) - f(\hat\theta^\ast)] &\le&  f(\theta^{k}) - f(\hat\theta^\ast) -\alpha \nabla f(\theta^{k})^\top \rmM\E \big[ \nabla \ell(\theta^k, x) \big] + \tfrac{\alpha^2}{2} \E \big[ \Tr \big(\rmM^\top \rmH \rmM \nabla \ell(\theta^k, x) \nabla \ell(\theta^k, x)^\top  \big) \big] \\
 &\leq&  f(\theta^{k}) - f(\hat\theta^\ast) -\alpha \nabla f(\theta^{k})^\top \rmM \nabla f(\theta^{k}) + \tfrac{\alpha^2}{2} \Tr \big(\rmM^\top \rmH \rmM (\rmC + \nabla f(\theta^k) \nabla f(\theta^k)) \big)  \\
 &=&  f(\theta^{k}) - f(\hat\theta^\ast) -\alpha \nabla f(\theta^{k})^\top (\rmM - \frac{\alpha}{2} \rmM^\top \rmH \rmM) \nabla f(\theta^{k}) + \tfrac{\alpha^2}{2} \Tr \big(\rmM^\top \rmH \rmM \rmC) \big)~,
\end{eqnarray*}
where in the second inequality we have used the covariance bound.

For $\mu_{M} \rmI \preccurlyeq \rmM - \frac{\alpha}{2} \rmM^\top \rmH \rmM$ and using the strong convexity bound $\tfrac{1}{2 \mu} \| \nabla f(\theta) \|^2 \ge f(\theta) - f(\hat\theta^\ast)$, we can simplify to 
\begin{eqnarray*}
\E [ f(\theta^{k+1}) - f(\hat\theta^\ast)] &\le&  f(\theta^{k}) - f(\hat\theta^\ast) -\alpha \mu_M \nabla f(\theta^{k})^\top \nabla f(\theta^{k}) + \tfrac{\alpha^2}{2} \Tr \big(\rmM^\top \rmH \rmM \rmC \big)  \\
&\le& f(\theta^{k}) - f(\hat\theta^\ast) -2 \alpha \mu_M \mu  \big( f(\theta^{k}) - f(\hat\theta^\ast)\big)  + \tfrac{\alpha^2}{2} \Tr \big(\rmM^\top \rmH \rmM \rmC \big)  \\
&=& \big(1 -2 \alpha \mu_M \mu\big)  \big( f(\theta^{k}) - f(\hat\theta^\ast)\big)  + \tfrac{\alpha^2}{2} \Tr \big(\rmM^\top \rmH \rmM \rmC \big)  \\
\end{eqnarray*}
Assuming $ \alpha \mu_M \mu \le \frac{1}{2}$, we have $\sum_{i=0}^k \big(1 -2 \alpha \mu_M \mu\big)^i \le \sum_{i=0}^\infty \big(1 -2 \alpha \mu_M \mu\big)^i = \frac{1}{2 \alpha \mu_M \mu} $. Therefore

\begin{eqnarray*}
\E [ f(\theta^{k+1}) - f(\hat\theta^\ast)] &\le&  f(\theta^{k}) - f(\hat\theta^\ast) -\alpha \mu_M \nabla f(\theta^{k})^\top \nabla f(\theta^{k}) + \tfrac{\alpha^2}{2} \Tr \big(\rmM^\top \rmH \rmM \rmC \big)  \\
&\le& f(\theta^{k}) - f(\hat\theta^\ast) -2 \alpha \mu_M \mu  \big( f(\theta^{k}) - f(\hat\theta^\ast)\big)  + \tfrac{\alpha^2}{2} \Tr \big(\rmM^\top \rmH \rmM \rmC \big)  \\
&=& \big(1 -2 \alpha \mu_M \mu\big)  \big( f(\theta^{k}) - f(\hat\theta^\ast)\big)  + \tfrac{\alpha^2}{2} \Tr \big(\rmM^\top \rmH \rmM \rmC \big)  \\
\end{eqnarray*}
Assuming $ \alpha \mu_M \mu \le \frac{1}{2}$, we have $\sum_{i=0}^k \big(1 -2 \alpha \mu_M \mu\big)^i \le \sum_{i=0}^\infty \big(1 -2 \alpha \mu_M \mu\big)^i = \frac{1}{2 \alpha \mu_M \mu} $. Taking full expectations and chaining inequalities we then have

\begin{eqnarray*}
\E [ f(\theta^{k}) - f(\hat\theta^\ast)] &\le& \big(1 -2 \alpha \mu_M \mu\big)^k  \big( f(\theta^{0}) - f(\hat\theta^\ast)\big)  + \tfrac{\alpha}{4 \mu_M \mu} \Tr \big(\rmM^\top \rmH \rmM \rmC \big)  \; .
\end{eqnarray*}
This concludes the proof.

\subsubsection{Convergence to limit cycles in the quadratic case}
\label{proof:quadratic}
For SGD with constant stepsize $\alpha$ and preconditioner $\rmM$, the update equation on the parameters is 
$$\theta_{t+1} = \theta_t - \alpha \rmM (\nabla f(\theta_t) + \epsilon_t) $$
In our quadratic case, $\nabla f(\theta_t) = \rmH (\theta_t - \theta^\ast)$ with $\E [\epsilon_t] = 0$ and $\E [\epsilon_t \epsilon_t^\top] = \rmS$. By defining $\delta_t = \E [\theta_t - \theta^\ast]$, we have

\begin{eqnarray*}
    \delta_{t+1} &=& (\rmI - \alpha \rmM \rmH) \delta_t \\
     &=&  (\rmI - \alpha \rmM \rmH)^{t+1} \delta_0
\end{eqnarray*}
This concludes the first result of proposition on the quadratic case.

By defining, $\rmSS_t = \E [ (\theta_t - \theta^\ast)  (\theta_t - \theta^\ast)^\top ]$, we get

\begin{eqnarray}
\rmSS_{t+1} &=& \rmSS_t - \E \big[ \alpha \rmM \big(\rmH (\theta_t - \theta^\ast) + \epsilon_t\big)  (\theta_t - \theta^\ast)^\top\big] \\ 
&-& \alpha \E \big[  (\theta_t - \theta^\ast) \big(\theta_t - \theta^\ast +\epsilon_t\big)^\top \rmH \rmM^\top\big]  \\
&+& \alpha^2\E \big[  \rmM \rmH (\theta_t - \theta^\ast) (\theta_t - \theta^\ast)^\top \rmH \rmM^\top \big] \\
&+& \alpha^2\E \big[  \rmM  \epsilon_t \epsilon_t^\top  \rmM^\top \big] \\
&=& \rmSS_t - \alpha \rmM \rmH \rmSS_t - \alpha \rmSS_t \rmH \rmM^\top + \alpha^2 \rmM \rmH \rmSS_t \rmH \rmM^\top + \alpha^2 \rmM \rmS \rmM^\top\\
&=& (\rmI - \alpha \rmM \rmH) \rmSS_t (\rmI - \alpha \rmM \rmH)^\top +  \alpha^2 \rmM \rmS \rmM^\top\\
\end{eqnarray}

\subsection{Expected suboptimality for SG and Polyak momentum on quadratic functions}
\label{sec:variance_computation}
We detail here the computation of the expected suboptimality at each timestep when optimizing a quadratic function with a diagonal Hessian when the noise is also diagonal. Note that all these results apply if $\rmH$ and $\rmS$ are simultaneously diagonalizable by a change of basis.

We assume that $f$ is a quadratic with Hessian $\rmH$ and that, at each time step, we receive a gradient perturbed by a random variable $\epsilon$ with $\E[\epsilon] = 0$, $\E[\epsilon \epsilon^\top] = \rmS$.
Further, we shall assume that $\rmH$ and $\rmS$ are both diagonal. With these assumptions, the optimization occurs in each dimension independently and we can thus focus on a single dimension. We will denote by $h$ and $c$ the hessian and noise variance along that direction.

\subsubsection{Proof of proposition~\ref{prop:sg_limit}}
We can compare this result to the same setting where we use stochastic gradient with a diagonal preconditioning matrix $\rmM$. Then we get
\begin{align*}
    s_i     &= (1 - \alpha \rmM_{ii}\rmH_{ii})^2 s_i + \alpha^2 \rmM_{ii}^2 \rmS_{ii}\\
    s_i     &= \frac{\alpha \rmM_{ii} \rmS_{ii}}{2 \rmH_{ii} - \alpha \rmM_{ii}\rmH_{ii}^2}\; ,
\end{align*}
and
\begin{align*}
    \E[f(\theta_t) - f(\hat\theta^\ast)]    &= \frac{1}{2} \sum_i\frac{\alpha \rmM_{ii} \rmS_{ii}}{2 - \alpha \rmM_{ii}\rmH_{ii}} + \gO(e^{-t}) \; .
\end{align*}

Generalizing to simultaneously diagonalizable matrices, we get
\begin{align*}
    \E[f(\theta_t) - f(\hat\theta^\ast)]    &= \frac{\alpha}{2} \Tr((2\rmI - \alpha \rmM\rmH)^{-1} \rmM\rmS) + \gO(e^{-t}) \; .
\end{align*}

\subsubsection{Proof of proposition~\ref{prop:polyak_limit}}
Polyak momentum update equations are:
\begin{align}
    v_t &= \gamma v_{t-1} + \nabla f(\theta_t) + \epsilon\\
    \theta_{t+1}    &= \theta_t - \alpha v_t \; .
\end{align}
Using the quadratic assumption, we can rewrite
\begin{align*}
    v_{t+1} &= \gamma v_t + \nabla f(\theta_{t+1}) + \epsilon\\
            &= \gamma v_t + h\theta_{t+1} + \epsilon\\
            &= \gamma v_t + h\theta_t - \alpha h v_t + \epsilon \; ,
\end{align*}
and the full update can be written in matrix form
\begin{equation}
\left[
\begin{array}{c}
\theta_{t}\\
v_{t}
\end{array}
\right]
=
\left[
\begin{array}{cc}
1   &   -\alpha\\
h   &   \gamma - \alpha h
\end{array}
\right]
\left[
\begin{array}{c}
\theta_{t-1}\\
v_{t-1}
\end{array}
\right]
+
\left[
\begin{array}{c}
0\\
\epsilon
\end{array}
\right]
\end{equation}

Denoting $\displaystyle P = \left[
\begin{array}{cc}
1   &   -\alpha\\
h   &   \gamma - \alpha h
\end{array}
\right]$ and $\displaystyle S_t = [\left[
\begin{array}{c}
\theta_{t}\\
v_{t}
\end{array}
\right]\left[
\begin{array}{c}
\theta_{t}\\
v_{t}
\end{array}
\right]^T$, we have
\begin{align}
    \E[S_t | S_{t-1}]
&= PS_{t-1} P^T + \left[
\begin{array}{cc}
0   &   0\\
0   &   c
\end{array}
\right] \; .
\end{align}

If there is a limit cycle for $\displaystyle \left[
\begin{array}{c}
\theta_{t}\\
v_{t}
\end{array}
\right]$, it will satisfy
\begin{align}
    S &= PSP^T + \left[
\begin{array}{cc}
0   &   0\\
0   &   c
\end{array}
\right] \; .
\end{align}

Writing $\displaystyle S = \left[
\begin{array}{cc}
s_\theta   &   s_{v\theta}\\
s_{v\theta}   &   s_v
\end{array}
\right]$, we have
\begin{align*}
    s_\theta &= s_\theta - 2\alpha s_{v\theta} + \alpha^2 s_v\\
    s_v &=  h^2s_\theta + 2 h(\gamma - \alpha h)s_{v\theta} + (\gamma - \alpha h)^2 s_v + c\\
    s_{v\theta} &= hs_\theta + (\gamma - 2\alpha h) s_{v\theta} - \alpha(\gamma - \alpha h) s_v \; .
\end{align*}
The first equation gives $s_{v\theta} = \frac{\alpha}{2}s_v$ and the last one becomes
\begin{align*}
    \frac{\alpha}{2}s_v &= hs_\theta + (\gamma - 2\alpha h) \frac{\alpha}{2}s_v - \alpha(\gamma - \alpha h) s_v\\
    s_\theta &= \frac{\alpha (1 + \gamma)}{2h} s_v \; .
\end{align*}
Finally, the second equation gives
\begin{align*}
    s_v &=  \left(h^2\frac{\alpha (1 + \gamma)}{2h} + 2 h(\gamma - \alpha h)\frac{\alpha}{2} + (\gamma - \alpha h)^2\right) s_v + c\\
    s_v &= \frac{c}{(1 - \gamma) \left(1 + \gamma - \frac{\alpha h}{2}\right)}
\end{align*}
and
\begin{align*}
    s_\theta &= \frac{\alpha(1 + \gamma) c}{h (1 - \gamma) (2 + 2\gamma - \alpha h)} \; .
\end{align*}
Adding all dimensions together and multiplying by the Hessian to get the value function, we get
\begin{align*}
    \E[f(\theta_t) - f(\hat\theta^\ast)]    &= \frac{1}{2} \sum_i\frac{\alpha(1 + \gamma) \rmS_{ii}}{(1 - \gamma) (2 + 2\gamma - \alpha \rmH_{ii})} + \gO(e^{-t}) \; .
\end{align*}

Generalizing to simultaneously diagonalizable matrices, we get
\begin{align}
    \E[f(\theta_t) - f(\hat\theta^\ast)]    &= \frac{\alpha}{2} \frac{(1 + \gamma)}{(1 - \gamma)}\Tr\left((2 (1 + \gamma)\rmI - \alpha \rmH)^{-1} \rmS\right) + \gO(e^{-t}) \; .
\end{align}

\subsubsection{Comparison between stochastic gradient and Polyak momentum in the large noise regime}

When the desired suboptimality is small, it requires a small $\alpha$ and the two suboptimality can be approximated by
\begin{align*}
    f(\theta_t) - f(\hat\theta^\ast)    &\approx \frac{1}{4} \sum_i\frac{\alpha \rmS_{ii}}{(1 - \gamma)} + o(1) \tag{Momentum}\\
    f(\theta_t) - f(\hat\theta^\ast)    &\approx \frac{1}{4} \sum_i\alpha \rmS_{ii} + o(1)\tag{Stochastic gradient} \; ,
\end{align*}
and we see that momentum needs a stepsize $\alpha$ that is $(1 - \gamma)$ times that of stochastic gradient to achieve the same suboptimality, countering any gain. This is what we see in Table~\ref{tab:stepsize_noise}.

\section{Experimental details}

\subsection{Details on the Hessian inverse}
\label{app:inversion}
As $\rmH$ is highly degenerate in neural networks, we compute an inverse of $\rmH$ by cutting all the eigenvalues smaller than $10^{-3} \times \lambda_{max}$ where $\lambda_{max}$ is the biggest eigenvalue of $\rmH$. We observed that $10^{-3}$ and $10^{-3}$ were reasonable constants for selecting the eigenvalues of significant magnitude. Using smaller constant sometimes lead to very noisy estimates of the TIC while using a bigger constant would lead to severe underestimation of the criterion.

\subsection{Details on the large scale experiments}
\label{app:exp_details}
These details apply for the experiments conducted in subsection~\ref{sec:neyshabur_gen}, figure~\ref{fig:large_scale_tic} and all figures in subsection~\ref{sec:exp_hfc}.

We remind the reader the setup.
\begin{itemize}
    \item 5 different architectures: logistic regression, a 1-hidden layer and 2-hidden layer fully connected network, and 2 small convolutional neural networks (CNNs, one with batch normalization~\citep{ioffe2015batch} and one without);
    \item 3 datasets: MNIST, CIFAR-10, SVHN;
    \item 3 learning rates: $10^{-2}$, $5\cdot 10^{-3}$, $10^{-3}$ using vanilla SGD with momentum $\mu = 0.9$;
    \item 2 batch sizes: $64$, $512$;
    \item 5 dataset sizes: 5k, 10k, 20k, 25k, 50k.
\end{itemize}
We train for 750k steps and compute our metrics every 75k steps.

\paragraph{Data preprocessing:} We choose to greyscale, resize to $7 \times 7$ pixels and normalize all the images in the 3 datasets used (CIFAR-10, MNIST and SVHN). This way, we can design architectures with a relatively low number of parameters.

\paragraph{Architectures:}

\begin{itemize}
    \item \texttt{mlp}: This one is a one hidden layer MLP. Input size is $7 \times 7 = 49$ and output size is $10$. The default number of hidden units is $70$. We use ReLU activations.
    \item \texttt{big\_mlp}: The architecture is the same as above but with one additional hidden layer.
    \item \texttt{logreg}: This is simple a $49 \times 10$ linear classifier.
    \item \texttt{cnn}: It is a small CNN with 3 layers. A first conv layer with kernel $3 \times 3$, $0$ padding and $15$ channels. The next layer has $20$ channels and same parameters. The last layer has $10$ channels and directly outputs the class scores.
    \item \texttt{cnn\_bn}: Same architecture as above, except for a spatial batch-norm after the second layer.
\end{itemize}

\subsection{Details on experiments of subsection~\ref{sec:neyshabur_gen}}
For these experiments we train one hidden layer MLPs on SVHN. Each points is computed by training three times with three different random seed until convergence.
In figure~\ref{fig:vary_h}, the labels are kept without corruption and we vary the hidden size layer by using $\{8, 10, 16, 20, 25, 30, 40, 50, 60, 70, 80, 100\}$ hidden units in the hidden layer.

In figure~\ref{fig:vary_corr}, we fix the number of hidden units to $70$ but we vary the labels corruption percentage from $0\%$ to $100\%$ (included) by increments of $10\%$.

The networks are trained for $150k$ gradients steps with a learning rate of $5e{-3}$ and a batch size of $256$.
We used a subset of $2000$ samples of SVHN to remain in the highly overparametrized regime, our networks were able to fit random data.

\begin{figure}[H]
  \centering
  \begin{subfigure}[b]{0.45\textwidth}
    \includegraphics[width=\textwidth]{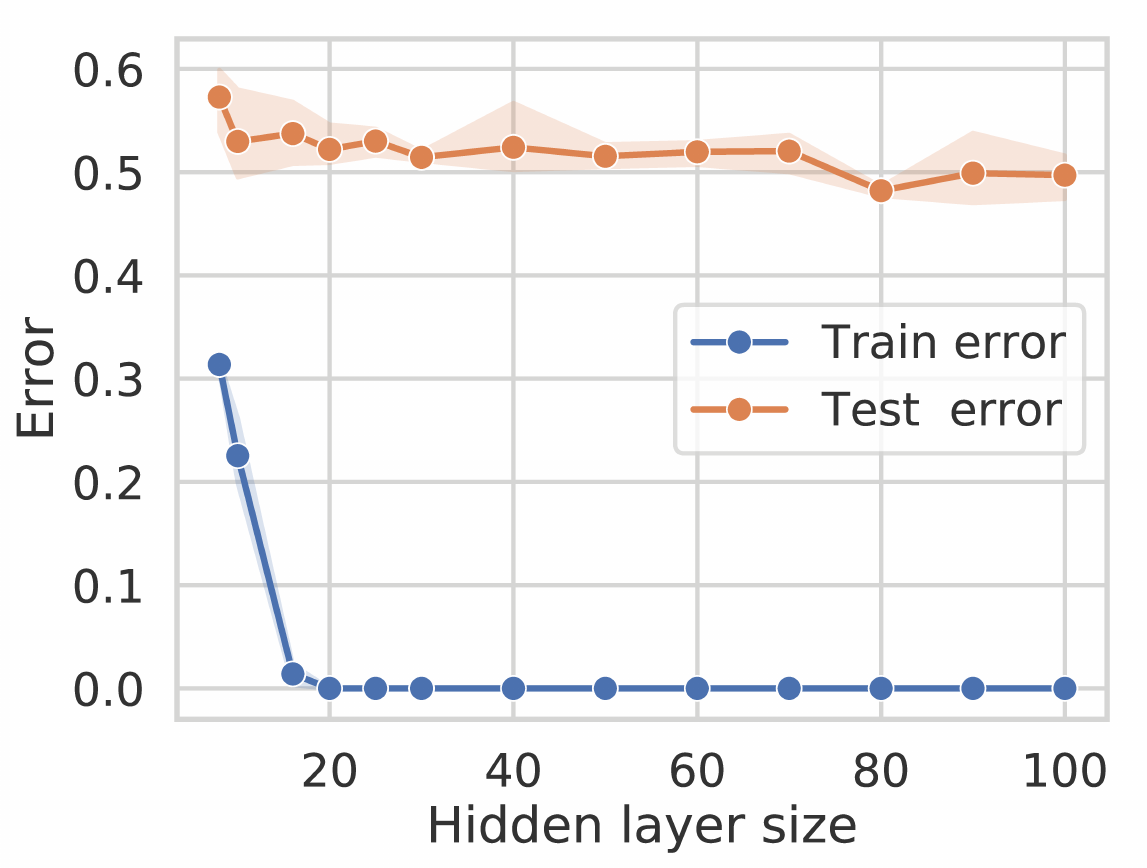}
    \caption{Varying hidden layer size.}
  \end{subfigure}
  \hfill
  \begin{subfigure}[b]{0.45\textwidth}
    \includegraphics[width=\textwidth]{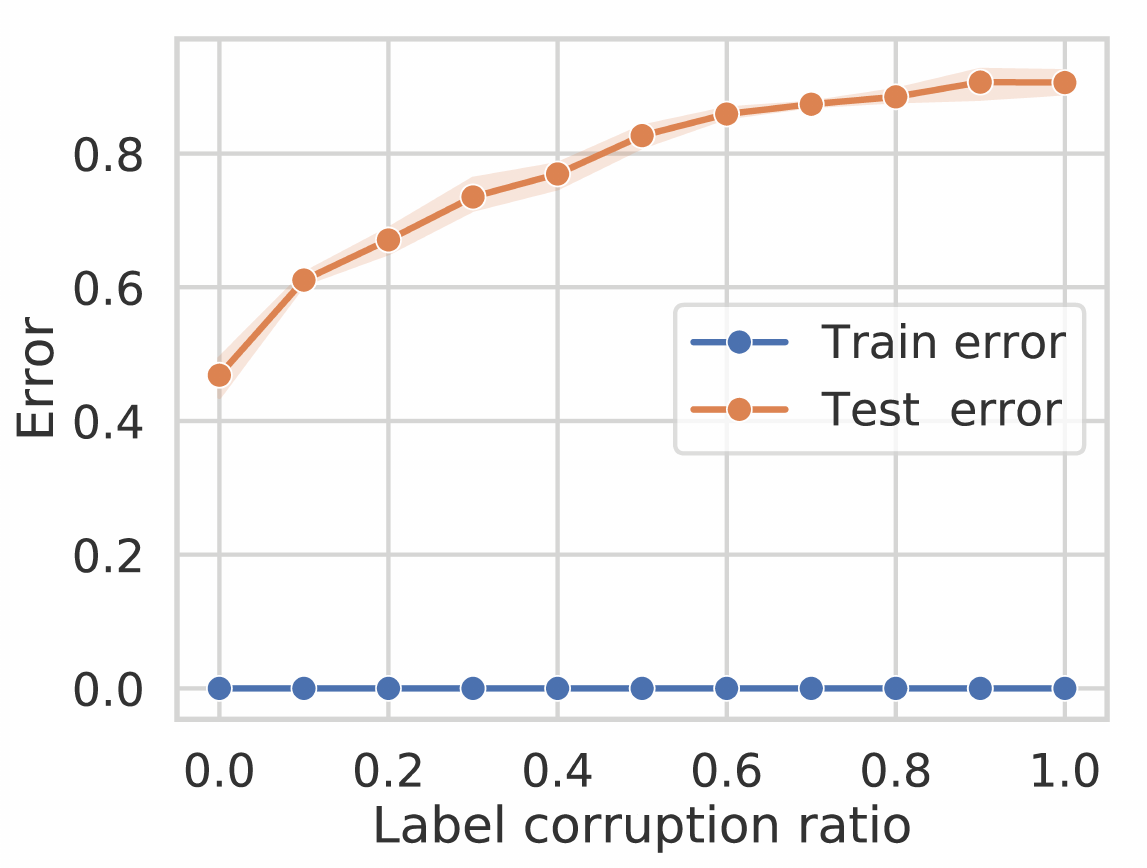}
    \caption{Varying label randomization level.}
  \end{subfigure}
\caption{The train and test errors associated with the experiments~\ref{fig:vary_h} and~\ref{fig:vary_corr}. We see that while we use small networks, they are still able to fit the data completely provided we use more than 20 hidden units. This behavior mirrors the one of bigger networks.}
\end{figure}

\end{document}

%% file: math_commands.tex
\usepackage{amsmath,amsfonts,bm}

\def\eqref#1{eq.~\ref{#1}}

\def\1{\bm{1}}

\def\rmSS{{\mathbf{\Sigma}}}
\def\rmA{{\mathbf{A}}}
\def\rmB{{\mathbf{B}}}
\def\rmC{{\mathbf{C}}}

\def\rmF{{\mathbf{F}}}

\def\rmH{{\mathbf{H}}}
\def\rmI{{\mathbf{I}}}

\def\rmM{{\mathbf{M}}}

\def\rmS{{\mathbf{S}}}

\DeclareMathAlphabet{\mathsfit}{\encodingdefault}{\sfdefault}{m}{sl}
\SetMathAlphabet{\mathsfit}{bold}{\encodingdefault}{\sfdefault}{bx}{n}

\def\gD{{\mathcal{D}}}

\def\gG{{\mathcal{G}}}

\def\gN{{\mathcal{N}}}
\def\gO{{\mathcal{O}}}

\def\gX{{\mathcal{X}}}
\def\gY{{\mathcal{Y}}}

\def\sR{{\mathbb{R}}}

\newcommand{\E}{\mathbb{E}}

\newcommand{\R}{\mathbb{R}}

\DeclareMathOperator{\Tr}{Tr}